\documentclass[journal,comsoc]{IEEEtran}
\usepackage{graphicx, color}
\usepackage{amssymb,amsmath,amsthm,amsfonts,bm,bbm,mathtools,booktabs,cite}
\usepackage{algorithm,algpseudocode}
\usepackage{optidef}
\usepackage{subfigure}
\usepackage[cmintegrals]{newtxmath}
\usepackage{ulem}

\newcommand{\argmax}{\mathop \mathrm{arg~max}\limits}
\renewcommand{\emph}[1]{\textit{#1}}
\newtheorem{theorem}{Theorem}						

\newtheorem{definition}[theorem]{Definition}
\begin{document}

\newcommand{\blue}[1]{\textcolor{blue}{#1}}
\newcommand{\pointer}[1]{
  \textcolor{blue}{\textbf{#1}}
}
\newcommand{\red}[1]{\textcolor{red}{#1}}

\algnewcommand\algorithmicexploration{\quad\textit{1. Exploring and Storing Experience:}}
\algnewcommand\Exploration{\item[\algorithmicexploration]}

\algnewcommand\algorithmictraining{\quad\textit{2. Training DQNs:}}
\algnewcommand\Training{\item[\algorithmictraining]}

\algnewcommand\algorithmictarget{\quad\textit{3. Updating Target DQNs:}}
\algnewcommand\Target{\item[\algorithmictarget]}

\algnewcommand\algorithmiccheck{\quad\textit{4. Checking Protagonist Performance:}}
\algnewcommand\PerformanceCheck{\item[\algorithmiccheck]}

\algnewcommand\algorithmiccheckad{\quad\textit{5. Checking Adversary Performance:}}
\algnewcommand\PerformanceCheckad{\item[\algorithmiccheckad]}

\title{\fontsize{21pt}{25pt}\selectfont Zero-Shot Adaptation for mmWave Beam-Tracking on\\
  Overhead Messenger Wires through\\ Robust Adversarial Reinforcement Learning}

\author{Masao~Shinzaki,~\IEEEmembership{Student Member,~IEEE,}
  Yusuke~Koda,~\IEEEmembership{Graduate Student Member,~IEEE,}
  Koji~Yamamoto,~\IEEEmembership{Senior Member,~IEEE,}
  Takayuki~Nishio,~\IEEEmembership{Senior Member,~IEEE,}
  Masahiro~Morikura,~\IEEEmembership{Member,~IEEE,}
  Yushi~Shirato,~\IEEEmembership{Member,~IEEE,}
  Daisei Uchida,~\IEEEmembership{Member,~IEEE,}
  and~Naoki~Kita,~\IEEEmembership{Member,~IEEE}

  \thanks{Masao~Shinzaki, Koji~Yamamoto, and Masahiro~Morikura are with the Graduate School of Informatics, Kyoto University, Kyoto 6068501 Japan. e-mail: \{shinzaki@imc.cce., kyamamot@, morikura@\}i.kyoto-u.ac.jp.}

  \thanks{Yusuke~Koda is with the Centre for Wireless Communications, University of Oulu, 90014, Finland, e-mail: Yusuke.Koda@oulu.fi.}

  \thanks{Takayuki~Nishio is with School of Engineering, Tokyo Institute of Technology, Ookayama, Meguro-ku, Tokyo, 158-0084, Japan, e-mail: nishio@ict.e.titech.ac.jp.}

  \thanks{D. Uchida, Y. Shirato, and N. Kita are with the NTT Access Network Service Systems Laboratories, NTT Corporation, Hikarinooka, Yokosuka 2390847 Japan.}}

\markboth{Journal of \LaTeX\ Class Files,~Vol.~00, No.~00, 0000~0000}%
{Shell \MakeLowercase{\textit{et al.}}: Bare Demo of IEEEtran.cls for IEEE Communications Society Journals}

\maketitle

\begin{abstract}
  Millimeter wave (mmWave) beam-tracking based on machine learning enables the development of accurate tracking policies while obviating the need to periodically solve beam-optimization problems.
  However, its applicability is still arguable when training-test gaps exist in terms of environmental parameters that affect the node dynamics.
  From this skeptical point of view, the contribution of this study is twofold.
  First, by considering an example scenario, we confirm that the training-test gap adversely affects the beam-tracking performance.
  More specifically, we consider nodes placed on overhead messenger wires, where the node dynamics are affected by several environmental parameters, e.g, the wire mass and tension.
  Although these are particular scenarios, they yield insight into the validation of the training-test gap problems.
  Second, we demonstrate the feasibility of \textit{zero-shot adaptation} as a solution, where a learning agent adapts to environmental parameters unseen during training.
  This is achieved by leveraging a robust adversarial reinforcement learning (RARL) technique, where such training-and-test gaps are regarded as disturbances by adversaries that are jointly trained with a legitimate beam-tracking agent.
  Numerical evaluations demonstrate that the beam-tracking policy learned via RARL can be applied to a wide range of environmental parameters without severely degrading the received power.
\end{abstract}

\begin{IEEEkeywords}
  mmWave communications, beam-tracking, robust adversarial reinforcement learning, zero-shot adaptation, overhead messenger wire.
\end{IEEEkeywords}
\IEEEpeerreviewmaketitle

\section{Introduction}
\label{sec:introduction}
\IEEEPARstart{W}{ireless} communication technologies in fifth-generation (5G) mobile networks provide multigigabit-per-second data rates, which fulfill the backhaul rate requirements \cite{niu2015survey, rangan2014millimeter}.
A key technology of 5G systems is millimeter-wave (mmWave) communications, which is advantageous because its broad spectral band increases the communication capacity \cite{cao2018capacity}.
In contrast to time- and cost-intensive optical fiber deployments, a mmWave wireless backhaul network has the advantages of high flexibility, cost efficiency, and rapid deployment of backhaul connections \cite{niu2015survey}.

However, signals transmitted in the mmWave frequency band experience larger path loss, which mandates the use of large antenna arrays and adaptive control of the array weights to ensure the main lobes of the antennas are pointing toward each other\cite{wei2014key}.
This adaptive control of the antenna array is required not only for initial access, but also during an operation in a mobile scenario or quasi-static scenarios where mmWave nodes are occasionally displaced owing to various perturbations.
The latter is termed beam-tracking, and in view of the large overhead involved in tracking, developing efficient beam-tracking methods has attracted considerable research interest, as is briefly discussed below.

\subsection{Related Work and Motivations}

\begin{table*}[!t]
  \caption{Comparison from Previous Works on Beam-Tracking}
  \begin{center}
    \begin{tabular}{c c c c c}
      \toprule
      Reference                                                                                                               & \begin{tabular}{c} 1) Periodically solving beam-optimization?
      \end{tabular} & \begin{tabular}{c} 2) Learning-Based? \end{tabular} & \begin{tabular}{c}\textbf{Addressing Training-Test Gap?} \end{tabular} & \begin{tabular}{c}
      \end{tabular} \\
      \midrule
      \cite{11ad, alexandropoulos2017position, zhang2016tracking, va2016beam, shaham2019fast, larew2019adaptive, lim2019beam} & Yes                       & No                        & --                                                    \\
      \cite{lin2019beamforming, elbir2019joint, elbir2019deep, wang2018mmwave, wang2019mmwave, klautau20185g}                 & No                        & Yes (SL)                  & No                                                    \\
      \cite{wang2019reinforcement,  mismar2019deep}                                                                           & No                        & Yes (RL)                  & No                                                    \\
      Previous version \cite{shinzaki2020deep, koda2020millimeter}                                                            & No                        & Yes  (RL)                 & No                                                    \\
      This paper                                                                                                              & No                        & Yes  (RL)                 & \textbf{Yes}                                          \\
      \bottomrule
    \end{tabular}
    \label{tab:previous_works_ver2}
  \end{center}
\end{table*}

Previous research addressed the beam-tracking problem mainly via the following two approaches: 1) periodically solving beam-optimization problems \cite{11ad, alexandropoulos2017position, zhang2016tracking, va2016beam, shaham2019fast, larew2019adaptive, lim2019beam}, 2) learning a beam-tracking policy beforehand  \cite{lin2019beamforming, elbir2019joint, elbir2019deep, wang2018mmwave, wang2019mmwave, klautau20185g, wang2019reinforcement,  mismar2019deep, shinzaki2020deep, koda2020millimeter}.
In the first approach, a mmWave node optimizes the array antenna weights  based on the estimated channels or angles of arrival (AoAs)/angles of departure (AoDs).
For example, in the IEEE 802.11ad standard \cite{11ad}, channels are surveyed by steering the transmitter/receiver beams; thus the array antenna weights are optimized.
The work in \cite{alexandropoulos2017position} estimated the AoAs/AoDs, and optimized the array antenna weights based on these estimations.
The works in \cite{zhang2016tracking, va2016beam, shaham2019fast, larew2019adaptive, lim2019beam} used filtering methods to estimate the AoAs/AoDs based on the surveyed channel and calculated the optimal antenna weights.
Although adaptive, this approach incurs computational overhead to periodically solve optimization problems that scale a number of antennas.

In the second approach, with the help of powerful machine-learning (ML) techniques approximating the input-output relationships, appropriate antenna weights or beam steering angles are learned beforehand.
Although the computational overhead increases during the training procedure, appropriate antenna weights or beam-steering angles can be outputted with fewer computations than solving the optimization problems thereafter.
Prior studies mainly leveraged supervised learning (SL), where the training data for optimal angles are given in advance \cite{lin2019beamforming, elbir2019joint, elbir2019deep, wang2018mmwave, wang2019mmwave, klautau20185g}.
Wang \textit{et al.} and Mismar \textit{et al.} \cite{wang2019reinforcement, mismar2019deep} leveraged reinforcement learning (RL) techniques, where a mmWave node learns appropriate beam steering angles from the received power without being given the optimal angles as training data.
Our previous studies relating to this work \cite{shinzaki2020deep, koda2020millimeter} were also categorized as focusing on RL-based beam tracking, where we considered the beam tracking of a mmWave node placed on an overhead messenger wire.

Although this second approach is attractive with respect to its ability to perform beam tracking, its applicability to the real environment is arguable because this approach may experience a training-test gap in terms of the environmental parameters that affect the node dynamics, which is the main focus of this study.
Generally, ML models perform worse as the gap between training and testing increases in terms of dataset distributions in SL or parameters that determine the state dynamics in RL.
Hence, this performance deterioration may naturally occur in the above learning-based beam tracking when a training-test gap exists.
To the best of our knowledge, this problem has been overlooked in the aforementioned prior studies pertaining to learning-based beam tracking.
Note that this is also the main difference from our prior studies \cite{shinzaki2020deep, koda2020millimeter}, where the training-test gap affecting node dynamics was not considered; in this sense, the contribution is different from these previous studies.
The differences between this work and the previous work are summarized in Table~\ref{tab:previous_works_ver2}.

\subsection{Contributions}
In view of the above problem, this work addresses the following two questions as the contribution of this study: 1) what would happen if a training-test gap were to exist in learning-based beam-tracking? 2) If this gap has a negative effect on beam tracking, how could we solve the problem caused by this training-test gap?
To address the first question, we confirm that the training-test gap causes the received power to deteriorate via numerical evaluation.
To address the second question, we first conceptualize \textit{zero-shot adaptation} as our objective.
Here, zero-shot adaptation implies that a learning agent exhibits feasible performance in test scenarios even when there are training-test gaps.
To realize zero-shot adaptation in mmWave beam-tracking, we applied robust adversarial reinforcement learning (RARL)\cite{pinto2017robust}, which is detailed below.
It should be noted that when examining these problems, we consider a particular scenario in which a mmWave node is placed on an overhead messenger wire, similar to our previous work (\cite{shinzaki2020deep} and \cite{koda2020millimeter}).
This is because this scenario is affected by several parameters that determine the node dynamics, such as the mass and tension of the wire, which facilitate the investigation of the training-test gap problem.
We believe that, although we are considering a particular scenario, this work provides firsthand insight regarding the overlooked problem caused by the training-test gap and provides the opportunity to rethink learning-based systems in the wireless communication research area.
In view of this, the contributions of this work are summarized as follows:
\begin{itemize}
  \item
        Given an example scenario in which a mmWave node is placed on an overhead messenger wire (see Fig.~\ref{fig:system_model}), through numerical evaluations, we confirm that the gap between the training and test scenarios deteriorates the beam-tracking performance.
        This debunks the importance of addressing training-test gaps to provide reliable mmWave links, and to the best of our knowledge, this perspective has not yet been reported in the literature.
  \item We demonstrate the feasibility of zero-shot adaptation of learning-based mmWave beam-tracking in the aforementioned scenarios.
        The key idea is to leverage RARL, wherein a beam-tracking agent is trained competitively to correspond to an intelligent adversary that attempts to cause beam misalignment by introducing additional wind disturbance.
        Through numerical evaluations, we show that even if the test scenarios were different from the training scenarios in terms of the parameters that affect the node dynamics, the proposed method prevents a drastic performance loss in terms of received power without adaptively fine-tuning the test scenario.
\end{itemize}

In a nutshell, our main scope is to validate the effectiveness of adding an intelligent adversary during the training; thereby confirming the concept of the zero-shot adaptation in the context of mmWave beam-tracking.
Thus, particularly in the evaluation in Section~\ref{sec:simulation}, we focus on the difference between the proposed method with the adversary and a baseline method without the adversary.
For this reason, delving into the problems commonly applied to the RL-based beam-tracking method with and without the adversary (e.g., delay for tracking and the accuracy of state acquisitions) is beyond the scope of this paper.
We believe that without addressing these problems, our evaluations sufficiently validate the aforementioned contributions.

The remainder of this paper is organized as follows.
In Section~\ref{sec:motivation}, we introduce zero-shot adaptation to overcome the training-test gap in RL problems and provide the motivation for solving this training-test gap in the mmWave beam-tracking problem.
In Section \ref{sec:beam_tracking}, we formulate the system model as an RL task.
In Section \ref{sec:adversarial_rl}, we explain the adversarial RL algorithm.
In Section \ref{sec:simulation}, we describe the simulation evaluation of the proposed beam-tracking policy.
Finally, we present our conclusions in Section \ref{sec:conclusion}.

\section{Definition and Motivation of Zero-Shot Adaptation}
\label{sec:motivation}

\subsection{Definition of Zero-Shot Adaptation}
We define the zero-shot adaptation problem considered in this study.
Let us consider a Markov decision process (MDP) $(\mathcal{S}, \mathcal{A}, r, p_{\bm \theta})$, where $\mathcal{S}$ and $\mathcal{A}$ denote the state and action space, respectively, $r:\mathcal{S}\times\mathcal{A}\to\mathbb{R}$ is the reward function, and $p_{\bm \theta}:\mathcal{S}\times\mathcal{A}\to \mathcal{S}$ is the state transition rule.
Note that the state transition rule is subject to the static parameter $\bm \theta$, which we refer to as the environmental parameters.
Therein, at each time step $k=1, 2, \dots, K$, a decision maker observes a state $s_k\in\mathcal{S}$ and determines an action $a_k\in\mathcal{A}$ according to a policy $\pi:\mathcal{S}\to\mathcal{A}$, and receives a reward $r_k$.
The objective of the decision-maker is to seek for a policy that maximizes the expected sum of rewards $\sum_{k = 1}^{K}\gamma^{k-1}r_k$, where $\gamma\in[0, 1]$ is the discount factor.
Given the aforementioned decision process, we define the zero-shot adaptation problem as follows:
\begin{definition}[Zero-shot adaptation]
  Let us consider two MDPs $\mathit{MDP}_{\mathrm{train}} = (\mathcal{S}, \mathcal{A}, r, p_{\bm \theta_{\mathrm{train}}})$ and $\mathit{MDP}_{\mathrm{test}} = (\mathcal{S}, \mathcal{A}, r, p_{\bm \theta_{\mathrm{test}}})$ are different in terms of the environmental parameters, that is, $\bm \theta_{\mathrm{train}}\neq \bm \theta_{\mathrm{test}}$.
  During the learning procedure of a policy, the decision-maker can act only in $\mathit{MDP}_{\mathrm{train}}$ and cannot access $\bm \theta_{\mathrm{test}}$.
  MAt the same time, the decision-maker or another third-party agent can ``manipulate'' training data, that is, they can replace several elements in the obtained state-action trajectory $(s_1, a_1, \dots, s_{K}, a_{K})$ with other values.
  Given this constraint, zero-shot adaptation is defined as finding a policy that maximizes the expected sum of rewards in $\mathit{MDP}_{\mathrm{test}}$ without any re-training of the policy.
\end{definition}
In a nutshell, the zero-shot adaptation considered in this study is to find the optimal policy in an unseen environment for a decision-maker with the tolerance of manipulating the state-action trajectory.
Hence, the problem reduces the manipulation of the state-action trajectory, that is, the training data, where the RARL is an effective approach, as demonstrated throughout this study.

Note that this problem is not necessarily identical to a ``zero-shot learning problem'' \cite{wang2019survey} in an SL context in terms of the way in which the problem regarding the training-test gap is overcome.
In the zero-shot learning as defined in \cite{wang2019survey}, an SL model is trained to enable the model to classify not only samples with a class label seen during training, but also those with unseen classes during training.
Rather than achieving this by manipulating the training data, it is accomplished with ``auxiliary knowledge,'' which indicates pre-obtained feature information involving class labels unseen in training\footnote{A well-known example is the problem of classifying an image of a ``zebra.'' Even if an image of the zebra is not contained in the training data, it would be possible to predict that it is an image of a zebra when it is known that a zebra has the appearance of a ``striped horse,'' and images labeled with ``horse'' and ``striped'' appear in the training dataset.}.
However, we also consider our problem to be ``zero-shot'' by focusing on the common aspect underlying both problems concerned with the training-test gap.

\subsection{Motivation of Zero-Shot Adaptation in mmWave Beam-Tracking}
As is shown in the subsequent sections, we consider the beam-tracking problem on an overhead messenger wire.
This scenario involves several environmental parameters, and among these, we select an overall mass $m$ and spring constant (i.e., wire tension) $k_0$ of the wire to which a mmWave node is attached.
These parameters are hardly measured precisely, particularly when a messenger wire is pre-installed.
Hence, to attach the mmWave nodes to such a pre-installed messenger wire, a beam-tracking policy should be trained without accessing these parameters.

One possible approach is to train the beam-tracking policy after attaching the nodes to a pre-installed messenger wire such that $\bm \theta_{\mathrm{traing}} \simeq \bm \theta_{\mathrm{test}}$.
However, this approach has the following two drawbacks, both of which can be solved via zero-shot adaptation.
First, in this approach, because it is necessary to train the beam-tracking policy after the attachment, supplying the connections is delayed.
However, if the beam-tracking policy could be pre-trained by simulations via zero-shot adaptation, the connection could be supplied immediately, which would be preferable for real deployments.
Second, even if the beam-tracking policy could be trained quickly using this approach, the aforementioned parameters gradually but certainly vary with time owing to the degradation of the wire over time.
This yields another test scenario $\bm \theta'_{\mathrm{test}}$, where $\bm \theta_{\mathrm{train}} \neq \bm \theta'_{\mathrm{test}}$.
Hence, to obtain robustness against the yielded training-test gap, it is worthwhile to consider zero-shot adaptation.

\begin{figure}[t]
  \centering
  \includegraphics[width=0.6\columnwidth]{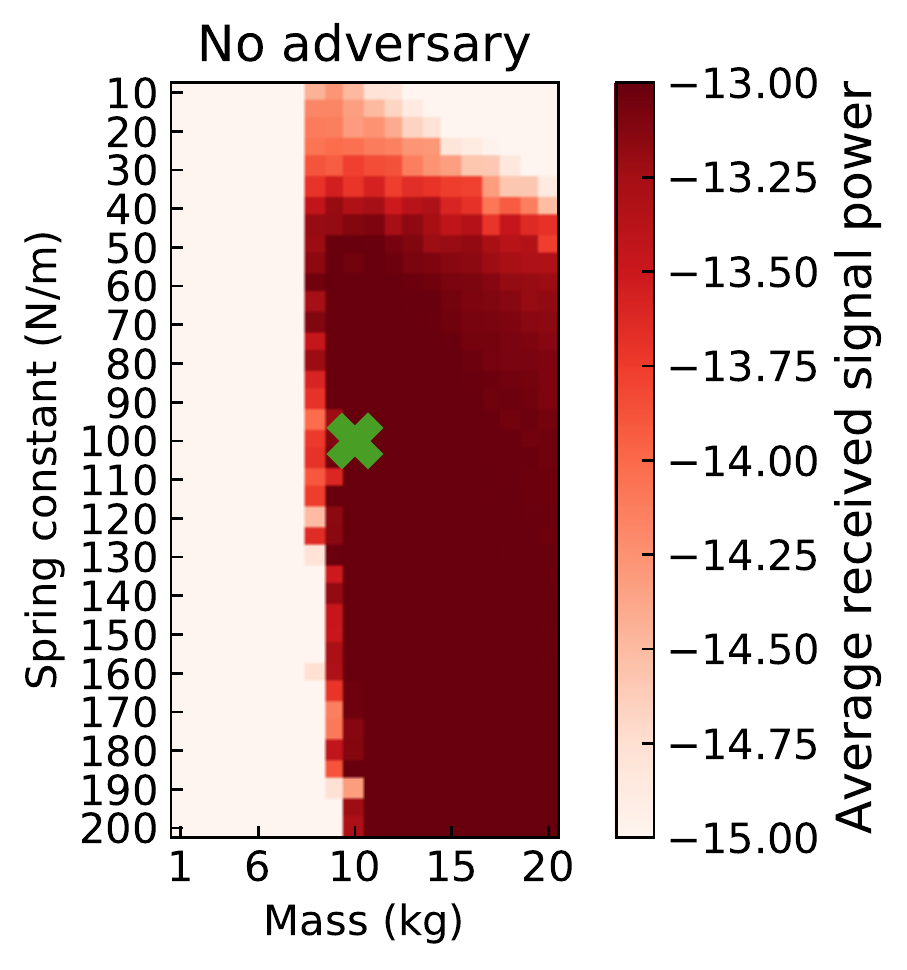}
  \caption{
  Heatmap depicting the robustness against training-test gap without zero-shot adaptation in terms of the received power.
  The cross mark point $\bm \theta_{\mathrm{train}} = [10\,\mathrm{kg}, 100\,\mathrm{N/m}]^{\mathrm{T}}$ represents the environmental parameters of the total wire mass and spring constant (i.e., wire tension) during training.
  The learned policy was brought in the different environmental parameter settings  $\bm \theta_{\mathrm{test}}$, where the wire mass and spring constant are depicted in horizontal and vertical axes, respectively.
  }
  \label{fig:noadv_result}
\end{figure}

At the same time, it remains unclear whether the training-test gap is harmful in the context of mmWave beam-tracking.
Hence, for the sake of clarity, we conclude this section by displaying a partial result in Section~\ref{subsec:simu_robustness}, as shown in Fig.~\ref{fig:noadv_result}, which shows what happens if there exists a training-test gap.
The beam-tracking policy was trained with the parameter setting indicated by a cross mark, that is, a wire mass of 10\,$\mathrm{kg}$ and wire tension 100\,$\mathrm{N/m}$; namely, $\bm \theta_{\mathrm{train}} = [10\,\mathrm{kg}, 100\,\mathrm{N/m}]^{\mathrm{T}}$.
We introduce this beam-tracking policy in other parameter settings $\bm \theta_{\mathrm{test}} = [m_{\mathrm{test}}, k_{\mathrm{test}}]^{\mathrm{T}}$, which are depicted as the horizontal and vertical axes in Fig.~\ref{fig:noadv_result}, respectively.
Hence, apart from the cross mark point, there exists a training-test gap, that is, $\bm \theta_{\mathrm{test}} \neq \bm \theta_{\mathrm{train}}$.
As shown in Fig.~\ref{fig:noadv_result}, under several settings of $\bm \theta_{\mathrm{test}}$, the learned beam-tracking policy does not perform better than under $\bm \theta_{\mathrm{train}}$ in terms of the received power.
In particular, the learned beam-tracking policy exhibited poorer received power when the wire mass was lower than that in the test scenario.
This is because a wire with a smaller mass has more vibrant wire dynamics, which makes beam-tracking more challenging than that in the training scenario.
This example led us to address the aforementioned zero-shot adaptation problem in the context of mmWave beam-tracking on a messenger wire.

\section{System Model}
\label{sec:beam_tracking}

\begin{figure}[!t]
  \centering
  \includegraphics[width=0.9\columnwidth]{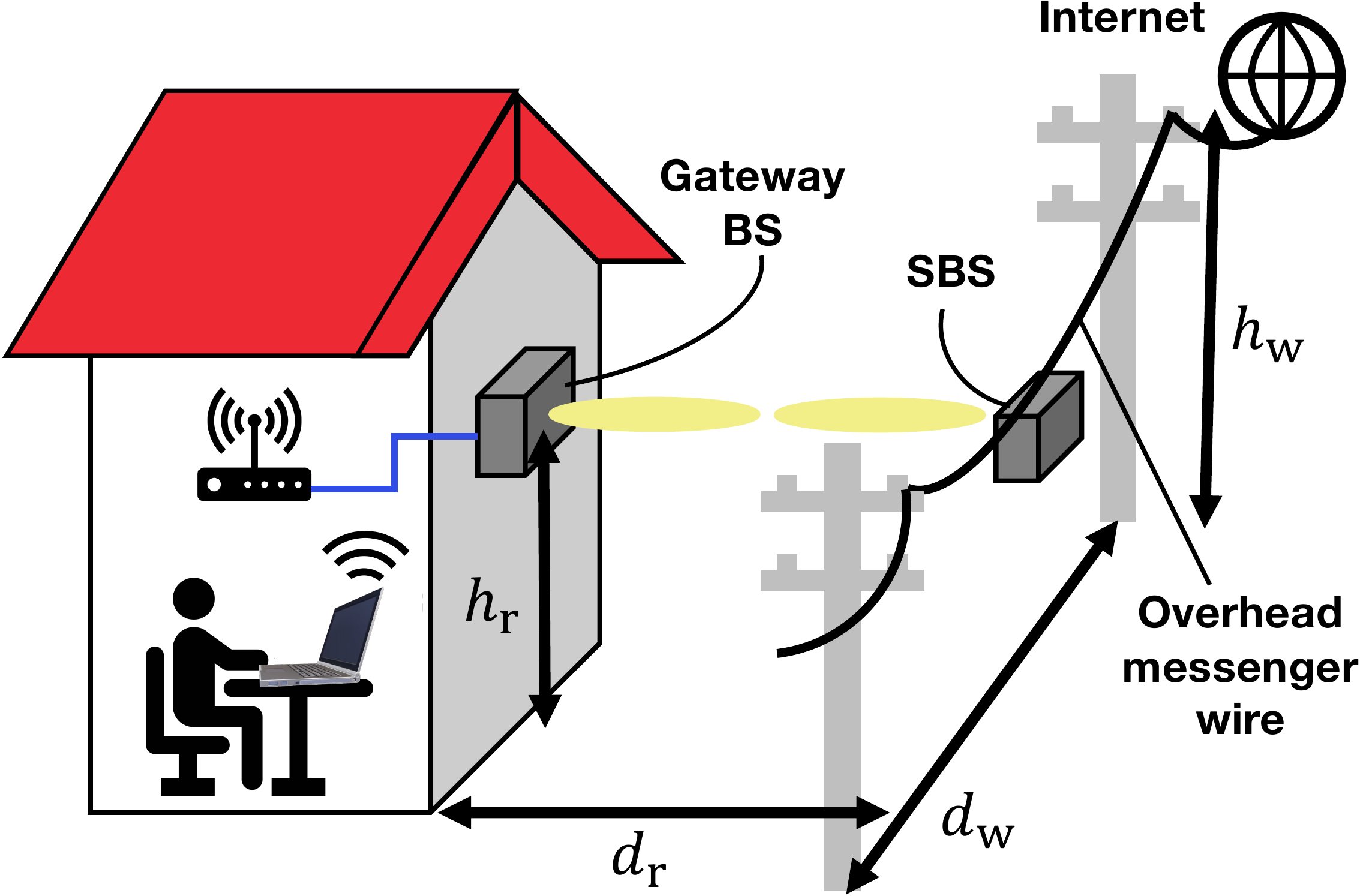}
  \caption{On-wire SBS in a millimeter-wave backhaul connection.}
  \label{fig:system_model}
\end{figure}

Fig.~\ref{fig:system_model} shows the system model for an on-wire small-cell base station (SBS) mmWave backhaul connection.
The SBS on the overhead messenger wire communicates with the gateway BS mounted on the building surface through a mmWave link to relay data from the gateway BS to the overhead messenger wire, and vice-versa.
The SBS is also connected to the optical fibers installed along the wire physically and receives the data to be transferred to the gateway BS via the upper layer protocol rather than the physical (PHY) layer (e.g., internet protocol).
A possible configuration for data delivery is as follows.
Both the SBS and gateway BS function as routers that forward the data from/to the end user, who sits in the house in Fig.~\ref{fig:system_model}.
The SBS hereby receives data routed to the end user via the upper layer protocol because the SBS is the natural waypoint for data delivery to the end user.
Briefly, the SBS provides a connection to the Internet infrastructure directly from an overhead messenger wire near the building.
However, it should be noted that the problem of beam-tracking that we consider in this study lies in the PHY layer, which is different from the data delivery problem in terms of the OSI reference model.
Hence, we focus on the former problem without a specific consideration for the upper layer protocols, which is sufficient to validate the effectiveness of the proposed beam-tracking method.
In this system, the SBS installed on an overhead messenger wire with a weight of $m$ performs beamforming to increase the received signal power of the gateway BS.
The endpoints of the overhead messenger wire are fixed to the telephone poles at a height $h_\mathrm w$, with a distance $d_\mathrm w$ between the poles.
The gateway BS is mounted at a height $h_\mathrm r$, and the distance between the overhead messenger wire and the gateway is $d_\mathrm r$.

For practical usage, beam-tracking on the side of both the gateway and the SBS should also be addressed; nonetheless, we consider only SBS-side beam tracking.
This is because of our focus on validating the effectiveness of adding the adversary during training, thereby confirming the feasibility of the zero-shot adaptation.
Indeed, gateway-side beam-tracking is more challenging than SBS-side beam-tracking in the sense that the gateway BS cannot immediately obtain the necessary state information in the RL-based beam tracking (i.e., position/velocity of the SBS).
However, this challenge is equally posed to beam-tracking trained both with and without an adversary, indicating that the difference between gateway-side and SBS-side beam tracking does not have a specific impact on the comparison between these two beam-tracking methods (i.e., with or without adversaries).
Hence, to focus on the comparison, we consider only SBS-side beam-tracking by assuming that the gateway-side beam-tracking was performed perfectly, which is sufficient to validate our contributions.


\begin{table}[t]
  \centering
  \caption{Summary of Notations in System Model}
  \begin{tabular}{cc}
    \toprule
    $h_\mathrm w$                                & Height of endpoints of wire                                                    \\
    $d_\mathrm w$                                & Direct distance between endpoints                                              \\
    $h_\mathrm r$                                & Height of gateway BS                                                           \\
    $d_\mathrm r$                                & Distance between wire and gateway BS                                           \\
    $m$                                          & Total wire mass                                                                \\
    $N$                                          & Number of proxy points on wire                                                 \\
    $i$                                          & Index of proxy point   on wire                                                 \\
    $k_0$                                        & Spring constant, i.e., coefficient for tensile force\cite{gladwell1986inverse} \\
    $\bm g$                                      & Gravitational acceleration                                                     \\
    $\bm a_i(t)$                                 & Acceleration of proxy point $i$ at time $t$                                    \\
    $\bm v_i(t)$                                 & Velocity of proxy point $i$ at time $t$                                        \\
    $\bm x_i(t)$                                 & Position of proxy point $i$ at time $t$                                        \\
    $c_0$                                        & Drag constant   \cite{zarate2016sde}                                           \\
    $\bm v_0(t)$                                 & Wind velocity                                                                  \\
    $\bm V_0$                                    & Covariance matrix of wind velocity                                             \\
    $\bm W_i(t)$                                 & Standard Wiener process                                                        \\
    $P_\mathrm t$                                & Transmit power                                                                 \\
    $\lambda$                                    & Radio-wave wavelength                                                          \\
    $\theta_{\mathrm{AoD}}, \phi_{\mathrm{AoD}}$ & Arbitrary zenith/azimuth angles                                                \\
    $\theta_{\mathrm{s}}, \phi_{\mathrm{s}}$     & Zenith/azimuth angles     of antenna main-lobe                                 \\
    $d$                                          & Distance between SBS and gateway BS                                            \\
    $A_\mathrm r$                                & Receiver antenna gain                                                          \\
    $G_{\mathrm{max}}$                           & Antenna gain of main-lobe                                                      \\
    $A_{\mathrm{m}}$                             & Front-back ratio \cite{rebato2018study}                                        \\
    $\theta_{\mathrm{3dB}}, \phi_{\mathrm{3dB}}$ & Horizontal/vertical 3\,dB beamwidth                                            \\
    $\mathit{SLA}_{\mathrm{V}}$                  & Side-lobe level limit \cite{rebato2018study}                                   \\
    $n_{\mathrm{V}}$, $n_{\mathrm{H}}$           & Number of vertical/horizontal array  elements                                  \\
    $\bm w$                                      & Beamforming vector                                                             \\
    $\Delta_\mathrm V$, $\Delta_\mathrm H$       & Vertical/horizontal array spacing distances                                    \\
    $\tau$                                       & Decision interval, i.e., interval between time steps                           \\
    $k$                                          & Index of time step                                                             \\
    $T$                                          & Observation time                                                               \\
    $\bm x^{(k)}_{\mathrm{S}}$                   & Position of SBS at time step $k$                                               \\
    $\bm x_{\mathrm{G}}$                         & Position of gateway BS                                                         \\
    \bottomrule
  \end{tabular}
  \label{tbl:parameters_definitions}
\end{table}

\subsection{Model of Dynamics in Overhead Messenger Wire}
\begin{figure}[t]
  \centering
  \includegraphics[width=0.7\columnwidth]{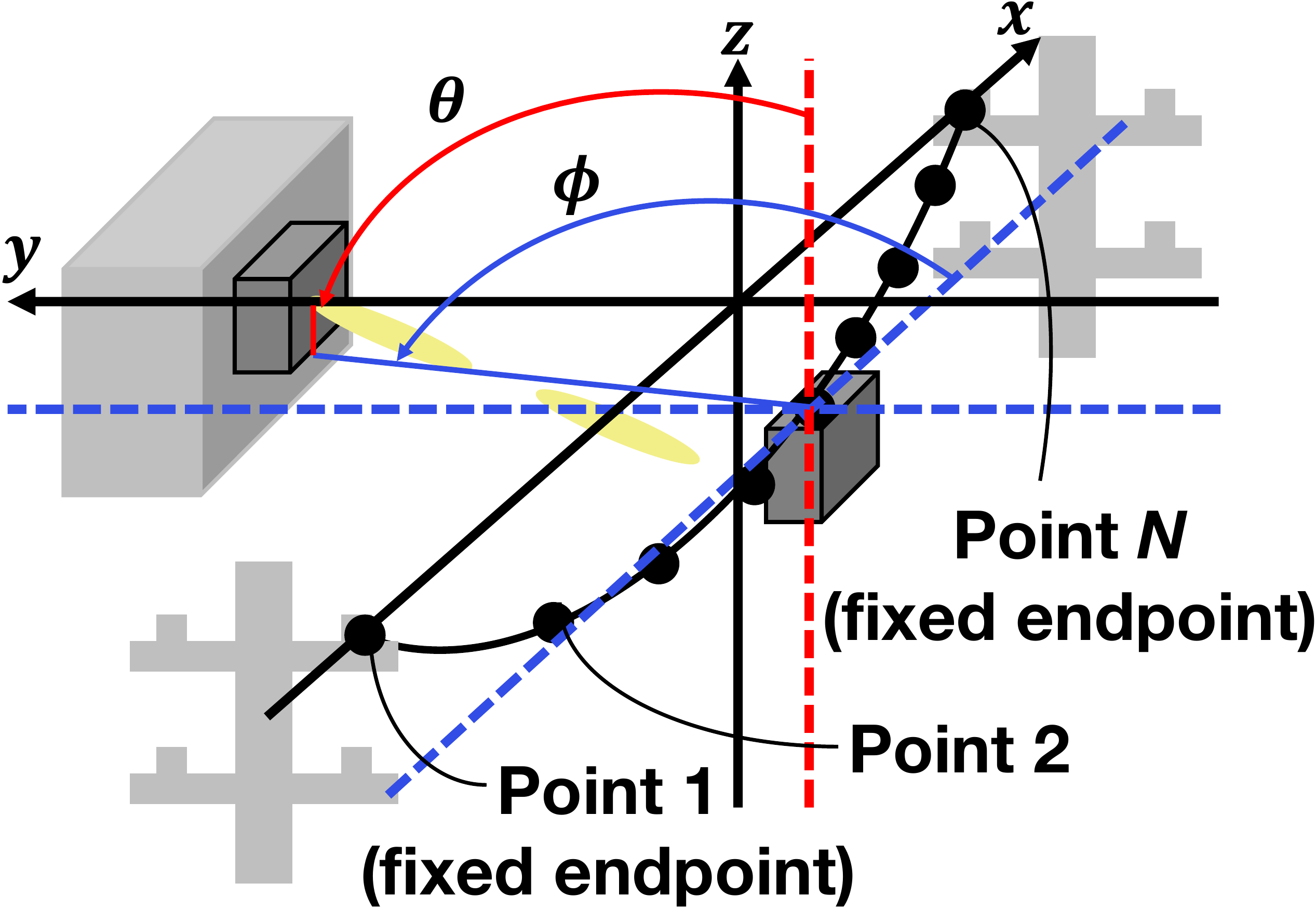}
  \caption{Coordinate system of the system model.}
  \label{fig:coordinate}
\end{figure}

According to \cite{gladwell1986inverse}, we modeled the overhead messenger wire as a chain of several proxy mass points that align and are separated by an equal distance, where the proxy mass points are affected by a tensile force from adjacent mass points.
Let $N$ and $m$ denote the number of proxy mass points and the total mass of the wire, respectively.
In the model, the mass of each point is assigned equally, that is, the mass of each point is $m/N$.
We denote points ~1 and ~$N$ as the ends of the chain of the mass points and term the residual mass points in the order of their proximity to point~1 as point~2,$\dots$, point~$N - 1$.
In the model, the tensile force applied to point~$i\in \{2, \dots, N - 1\}$ is proportional to the relative position of the adjacent points, i.e., points $i-1$ and $i+1$, where the total tensile force is calculated as:
$k_0\bigl[\bm x_{i+1}(t) + \bm x_{i-1}(t) - 2\bm x_i(t)\bigr]$, where $\bm x_i(t)$ is the position of point~$i$ at time $t$ measured in the coordinate system in Fig.~\ref{fig:coordinate}.
The term $k_0$ is constant and determines a wire tension.
This model can be regarded as a spring chain, where the mass points are connected via springs; hence, we refer to the constant $k_0$ as the ``spring constant,'' hereinafter.

We delve into the dynamics of these proxy mass points.
Let $\bm a_i(t)\in\mathbb R^3$ denote the acceleration of point $i$ at time $t$.
The accelerations of points $1$ and $N$ fixed to the telephone poles are expressed as $\bm a_1(t)=\bm a_N(t)=\bm 0$.
For $i=2,3,\dots,N-1$, from the equation of motion, $\bm a_i(t)$ is given by
\begin{align}
  \label{eq:accel}
  \bm a_i(t) = \bm g + \underbrace{\frac{k_0N}{m}\bigl[\bm x_{i+1}(t) + \bm x_{i-1}(t) - 2\bm x_i(t)\bigr]}_{\text{acceleration from tensile force}},
\end{align}
where $\bm g\in\mathbb R^3$ denotes the gravitational acceleration.

As the perturbations that are responsible for the dynamics in the wire, we consider a wind perturbation and consider the wind drag model in \cite{zarate2016sde}.
In this model, the wind drag primarily consists of frictional drag and pressure drag \cite{zarate2016sde}.
The frictional drag increases proportionally with the velocity of the mass point of interest $\bm v_i(t)$ relative to the wind velocity $\bm v_{\mathrm{o}}(t)$.
We denote the constant of the proportionality as $c_0$ and refer to it as the ``drag constant.''
The pressure drag has random magnitude regardless of the wind speed.
The derivatives of the velocity and position of point~$i$ at time $t$ are denoted by $\mathrm d\bm v_i(t)\in\mathbb R^3$ and $\mathrm d\bm x_i(t)\in\mathbb R^3$, respectively.
The derivatives of the velocities and positions of points $1$ and $N$ are expressed as $\mathrm d\bm v_1(t)=\mathrm d\bm v_N(t)=\mathrm d\bm x_1(t)=\mathrm d\bm x_N(t)=\bm 0$, respectively.
For $i=2,3,\dots,N-1$, $\mathrm d\bm v_i(t)$ and $\mathrm d\bm x_i(t)$ are calculated as follows \cite{zarate2016sde, shiri2019massive}:
\begin{align}
  \label{eq:pos_vel}
  \mathrm d\bm v_i(t) & = \bm a_i(t) \,\mathrm dt \underbrace{-c_0\bigl[\bm v_i(t)-\bm v_\mathrm o(t)\bigr] \mathrm dt + \bm V_\mathrm o \,\mathrm d\bm W_i(t)}_{\text{derivatives from the wind}}, \notag \\
  \mathrm d\bm x_i(t) & = \bm v_i(t) \,\mathrm dt,
\end{align}
where $\bm V_\mathrm o\in\mathbb R^{3\times 3}$ and $\bm W_i(t)\in\mathbb R^3$ denote the covariance matrix of the wind speed and the standard Wiener process that is independently and identically distributed across positions of point $i$, respectively.

\subsection{Radio-Wave Propagation Model}
According to the free-space path loss model \cite{friis1946note}, we consider the received signal power of the gateway BS to be determined by the distance $d$ between the SBS and the gateway BS and the antenna radiation pattern.
This free-space path loss model is consistent with the mmWave channel measurement conducted under a line-of-sight (LoS) conditions in an open space \cite{geng2008millimeter}, and this is a feasible assumption considering that the SBS and gateway BS in the above-mentioned scenario are likely to be deployed under such conditions.
Note that in a NLoS condition, the above assumption does not hold indeed; however, we do not consider the NLoS condition because our focus is on the beam-tracking inaccuracy caused by the training-test gap, which occurs even under LoS conditions.
In other words, considering only the LoS condition is sufficient to validate the contributions of this study.
We consider the use of a directional antenna; hence, the antenna radiation pattern is determined by AoDs in the zenith and azimuth angles. That is, $\theta_{\mathrm{AoD}}, \phi_{\mathrm{AoD}}$, respectively, and the zenith and azimuth steering angles, that is, $\theta_\mathrm s, \phi_\mathrm s$, respectively.
Note that these angles are measured in the coordinate system illustrated in Fig.~\ref{fig:coordinate}.
From the Friis transmission equation \cite{friis1946note}, $P_\mathrm r(d,\theta_{\mathrm{AoD}},\phi_{\mathrm{AoD}},\theta_\mathrm s,\phi_\mathrm s)$ is given by
\begin{align}
  P_\mathrm r(d,\theta_{\mathrm{AoD}},\phi_{\mathrm{AoD}},\theta_\mathrm s,\phi_\mathrm s) = \left(\frac{\lambda}{4\pi d}\right)^{\!\!2} A_\mathrm t(\theta_{\mathrm{AoD}},\phi_{\mathrm{AoD}},\theta_\mathrm s,\phi_\mathrm s)\, A_\mathrm rP_\mathrm{t},
\end{align}
where $P_\mathrm t$, $\lambda$, and $ A_\mathrm r$ are constants, and denote the transmission power of the SBS, wavelength of the radio waves, and receiver antenna gain, respectively.
Moreover, $A_\mathrm t(\theta_{\mathrm{AoD}},\phi_{\mathrm{AoD}},\theta_\mathrm s,\phi_\mathrm s)$ denotes the transmission antenna gain, with its maximum value at $\theta_{\mathrm{AoD}}=\theta_\mathrm s$ and $\phi_{\mathrm{AoD}}=\phi_\mathrm s$.
For the sake of simplicity, we omit the subscript $_{\mathrm{AoD}}$, hereinafter.

We considered the use of the array antenna model in \cite{3gpp,rebato2018study}, where the transmission antenna gain $A_\mathrm t(\theta,\phi,\theta_\mathrm s,\phi_\mathrm s)$ is given by
\begin{align}
  \label{eq:antenna_gain}
  A_\mathrm t(\theta, \phi,\theta_\mathrm s,\phi_\mathrm s) = A_\mathrm E(\theta, \phi) + \mathit{AF}(\theta, \phi, \theta_\mathrm s, \phi_\mathrm s),
\end{align}
where $A_\mathrm E(\theta,\phi)$ and $\mathit{AF}(\theta, \phi, \theta_\mathrm s, \phi_\mathrm s)$ denote the element radiation pattern and array factor, respectively.
The element radiation pattern $A_\mathrm E(\theta,\phi)$ of each single antenna element is composed of horizontal and vertical radiation patterns.
The element radiation pattern $A_\mathrm E(\theta,\phi)$ is given by
\begin{align}
  A_\mathrm E(\theta, \phi) = G_\mathrm{max} -\min\left\{-\left[A_{\mathrm E,\mathrm V}(\theta)+A_{\mathrm E,\mathrm H}(\phi)\right], A_\mathrm m\right\},
\end{align}
where $A_{\mathrm E,\mathrm V}(\theta)$, $A_{\mathrm E,\mathrm H}(\phi)$, $G_\mathrm{max}$, and $A_\mathrm m$ denote the vertical and horizontal radiation patterns, maximum directional gain of the antenna element, and front-back ratio, respectively.
The vertical and horizontal radiation patterns $A_{\mathrm E,\mathrm V}(\theta)$ and $A_{\mathrm E,\mathrm H}(\phi)$, respectively, are obtained as follows:
\begin{align}
  A_{\mathrm E,\mathrm V}(\theta) & = -\min\left\{12\left(\frac{\theta-90^\circ}{\theta_{3\mathrm{dB}}}\right)^{\!\!2}, \mathit{SLA}_\mathrm V\right\}, \notag \\
  A_{\mathrm E,\mathrm H}(\phi)   & = -\min\left\{12\left(\frac{\phi}{\phi_{3\mathrm{dB}}}\right)^{\!\!2}, A_\mathrm m\right\},
\end{align}
where $\theta_{3\mathrm{dB}}$, $\phi_{3\mathrm{dB}}$, and $\mathit{SLA}_\mathrm V$ are the vertical $3\,\mathrm{dB}$ beamwidth, horizontal $3\,\mathrm{dB}$ beamwidth, and side-lobe level limit, respectively.

The array factor $\mathit{AF}(\theta,\phi,\theta_\mathrm s,\phi_\mathrm s)$ models the directivity of the antenna array, which
is expressed for an array of $n=n_\mathrm V n_\mathrm H$ elements as
\begin{align}
  \mathit{AF}(\theta, \phi, \theta_\mathrm s, \phi_\mathrm s) = 10\log_{10}\Bigl[1+\Bigl(\left|(1/\sqrt{n})\bm w\right|^{2}-1\Bigr)\Bigr],
\end{align}
where $n_\mathrm V$ and $n_\mathrm H$ denote the number of vertical and horizontal elements, respectively, and $\bm w\in\mathbb C^n$ denotes the beamforming vector, which is given by
\begin{align}
  \bm w   & = \left[w_{1,1}, w_{1,2}, \ldots, w_{n_\mathrm V, n_\mathrm H}\right]^{\mathrm{T}}, \notag                     \\
  w_{p,r} & = \mathrm e^{\mathrm j2\pi[(p-1)\Delta_\mathrm V\Psi_p/\lambda + (r-1)\Delta_\mathrm H\Psi_r]/\lambda}, \notag \\
  \Psi_p  & = \cos\theta - \cos\theta_\mathrm s, \notag                                                                    \\
  \Psi_r  & = \sin\theta\sin\phi - \sin\theta_\mathrm s\sin\phi_\mathrm s,
\end{align}
where $\Delta_\mathrm V$ and $\Delta_\mathrm H$ denote the spacing distances between the vertical and horizontal elements of the array, respectively.
As a specific characteristic of mmWave communications, the spacing distances $\Delta_{\mathrm{V}}$ and $\Delta_{\mathrm{H}}$ are of the order of several millimeters (e.g.,  $\Delta_{\mathrm{V}}=\Delta_{\mathrm{H}}=2.5\,\mathrm{mm}$ at an the RF frequency of $60\,\mathrm{GHz}$).
This is because of the common setting where spacing distances should not exceed the half-wavelength to avoid high grating lobes.

The SBS periodically observes both the instantaneous received signal power and its position and velocity.
Hereinafter, we let the notation $\tau$ denote the observation interval and term the time instants for the observation as the ``time step.''
Accordingly, we consider the steering angle capable of moving up, down, left, or right by an angle $\beta$ at each time step.
The problem for determining the steering angles at each time step is formulated in the next section.

\subsection{Initial Access Procedure}
Among the initial access procedures, we only note the beam alignment between the SBS and gateway BS in the initial stage.
Indeed, many other procedures are mandatory to initialize the mmWave communications (e.g., device discovery and association frame exchanges); however, our focus is on beam-tracking, which is disjoint these initial access procedures.
Hence, providing a concrete design of the initial access procedures is basically beyond the scope of this paper, and we describe only the  beam alignment in the initial stage, which is the most relevant procedure for beam-tracking.

The beam alignment in the initial stage should allow the beam of the SBS to point toward the gateway BS; thereby allowing the gateway BS to receive the maximum received power.
Indeed, one can assume arbitrary beam-alignment procedures as long as the beam of the SBS to point toward the gateway BS.
As an example of this procedure, in the evaluation in Section~V, we employed position-aware beam alignment based on our assumption that the positions of the SBS and gateway BS are given in advance.
Therein, we simply calculated the orientation of the gateway BS from these positions under the initial conditions without wind.
Subsequently, we established the array antenna weights such that the beam pointed toward the gateway BS.

\subsection{Formulation}
Let $\mathcal N=\{1,2,\dots,\lfloor T/\tau\rfloor\}$ denote the set of indices of the time steps, where the index $k\in\mathcal{N}$ corresponds to the time $t = k\tau$.
The term $T$ is the total time length for the beam-tracking.
Moreover, we let the superscript $(k)$ indicate that the variables of interest are measured at the time $t = k\tau$.
The optimization problem is formulated as follows:
\begin{maxi!}[1]
{\substack{\left(a^{(k)}_\theta, a^{(k)}_\phi\right)_{k\in\mathcal{N}}}}
{\frac{1}{\lfloor T/\tau \rfloor}\sum_{k \in \mathcal{N}} P_\mathrm{r}\left(d^{(k)}, \theta^{(k)}, \phi^{(k)}, \theta_\mathrm s^{(k)}, \phi_\mathrm s^{(k)}\right)}{}{}.
\addConstraint{d^{(k)}}{= \left\|\bm x^{(k)}_\mathrm{S}-\bm x_\mathrm{G}\right\|}
\addConstraint{\theta^{(k)}}{= \arccos \frac{x^{(k)}_{\mathrm{S},\mathrm z}-x_{\mathrm{G}, \mathrm z}}{d^{(k)}}}
\addConstraint{\phi^{(k)}}{= \arctan \frac{x^{(k)}_{\mathrm{S},\mathrm y}-x_{\mathrm{G}, \mathrm y}}{x^{(k)}_{\mathrm{S}, \mathrm x}-x_{\mathrm{G}, \mathrm x}}}
\addConstraint{\theta^{(k)}_\mathrm s}{= \theta^{(k-1)}_\mathrm s + a^{(k)}_\theta\beta}
\addConstraint{\phi^{(k)}_\mathrm s}{= \phi^{(k-1)}_\mathrm s + a^{(k)}_\phi\beta}
\addConstraint{a^{(k)}_\theta, a^{(k)}_\phi}{\in \{-1, 0, 1\}}
\addConstraint{\left|a^{(k)}_\theta\right|+\left|a^{(k)}_\phi\right|}{\leq 1},
\end{maxi!}

\noindent where $a^{(k)}_\theta$ and $a^{(k)}_\phi$ denote the action for the zenith and azimuth angles at time step $k$, respectively, which can move the zenith and azimuth steering angles $\theta_\mathrm s^{(k)}, \phi_\mathrm s^{(k)}$ by angle $\beta$, respectively.
Moreover, $d^{(k)}$, $\theta^{(k)}$, and $\phi^{(k)}$ denote the distance, beam zenith, and azimuth angle from the SBS to the gateway BS at time step $k$, respectively.
These variables are defined by the positions of the SBS and gateway BS, i.e., $\bm x^{(k)}_\mathrm{S} = [x^{(k)}_{\mathrm{S}, \mathrm x}, x^{(k)}_{\mathrm{S}, \mathrm y}, x^{(k)}_{\mathrm{S}, \mathrm z}]^{\mathrm{T}} \in\mathbb R^3, \bm x_\mathrm{G} = [x_{\mathrm{G}, \mathrm x}, x_{\mathrm{G}, \mathrm y}, x_{\mathrm{G}, \mathrm z}]^{\mathrm{T}} \in\mathbb R^3$, respectively.

\section{Adversarial RL-based Beam-Tracking Based on Zero-Shot Adaptation}
\label{sec:adversarial_rl}

\subsection{Reason for Adversarial RL}
Motivated by the importance of zero-shot adaptation as discussed in Section~\ref{sec:motivation}, we propose a RARL-based beam-tracking method.
The key reason for using RARL is to develop the capability to overcome the training and test gap by: 1) regarding the gap as a disturbance from an adversarial agent that impedes the legitimate agent; 2) training both the adversarial and legitimate agent, thereby allowing the legitimate agent to experience more severe disturbances.

The explanation more specific to our beam-tracking problem is as follows: By training the adversarial agent to disturb the on-wire SBS with additional wind, the beam-tracking agent experiences more rapid displacements in the on-wire SBS than without such adversarial agents.
We hypothesize that this well simulates a situation in which it would be difficult to correct the directional beams, where the actual wire mass or spring constant is smaller than that used for  training.
This means that the adversarial agent provides richer experiences to the beam-tracking agent in view of the existence of the training and test gap; hence, the beam-tracking agent would be expected to obtain a robust beam-tracking policy against these training and test gaps.

\subsection{Overview of RARL-Based Beam-Tracking}
\begin{figure}[t]
  \centering
  \includegraphics[width=0.9\columnwidth]{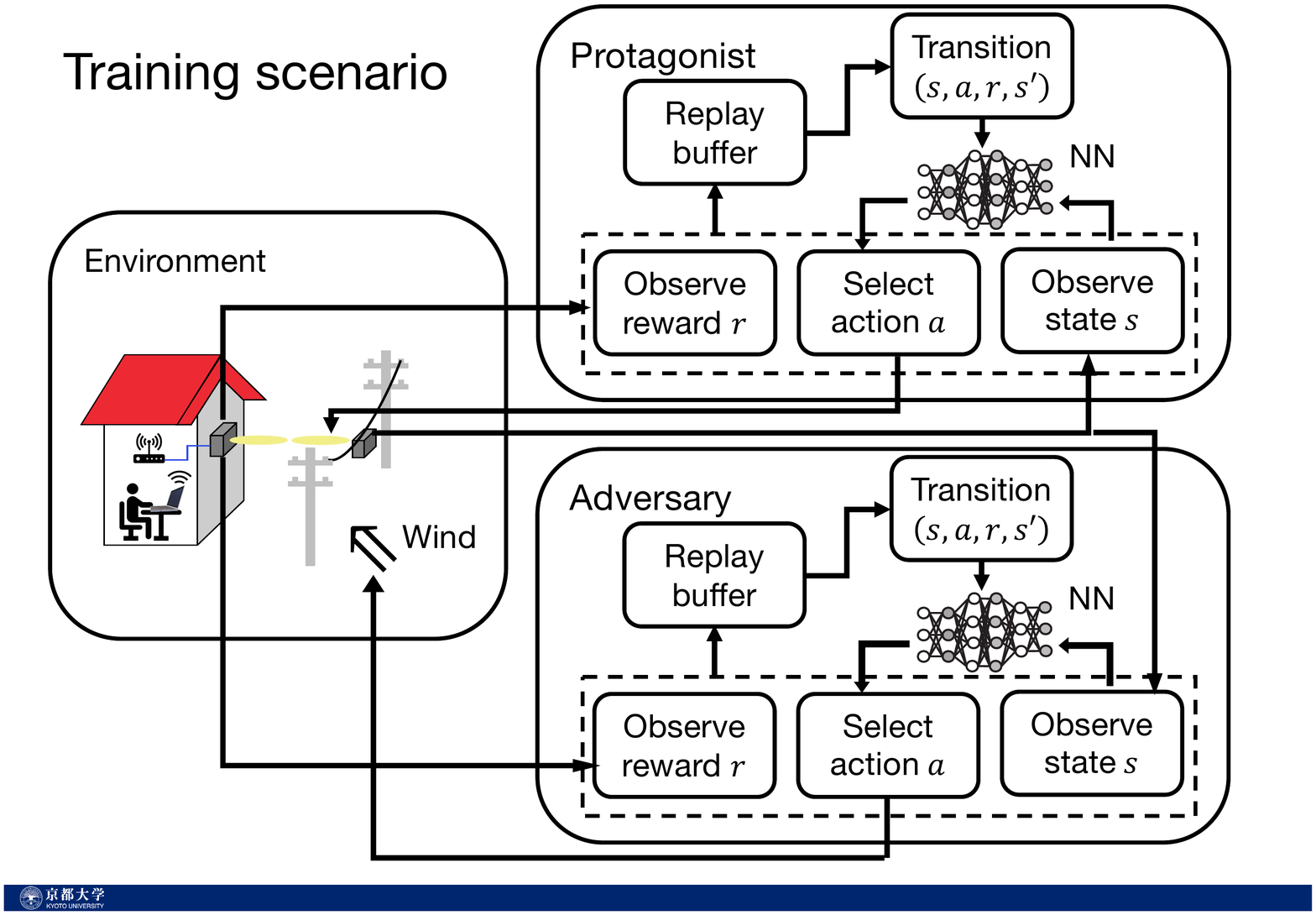}
  \caption{Training scenario of the adversarial RL.}
  \label{fig:scenario_train}
\end{figure}

\begin{figure}[t]
  \centering
  \includegraphics[width=0.9\columnwidth]{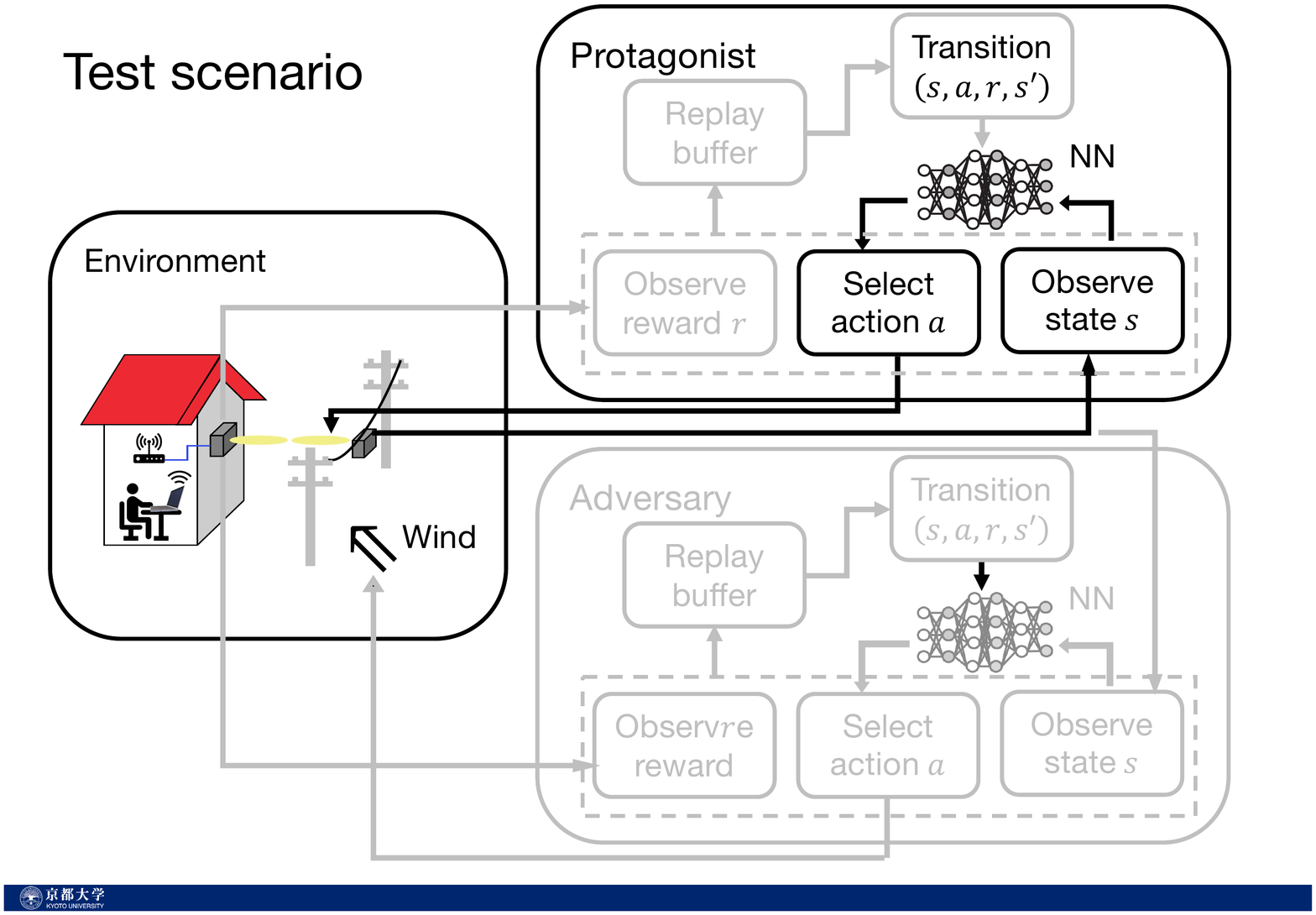}
  \caption{Test scenario of the adversarial RL.}
  \label{fig:scenario_test}
\end{figure}

The training procedure for the RARL-based beam-tracking is shown in Fig.~\ref{fig:scenario_train}.
In the training scenario, as shown in Fig.~\ref{fig:scenario_train}, the protagonist corresponding to the beam-tracking agent learns to maximize the average received signal power.
In contrast, the adversary learns to minimize the average received signal power by generating additional wind.
To achieve these purposes, the protagonist and adversary observe a state, select an action, and observe a reward to update their NN from experienced transitions.
In the test scenario shown in Fig.~\ref{fig:scenario_test}, the protagonist corrects the beam misalignment according to the policy learned in the training scenario.
To examine the feasibility of zero-shot adaptation, the environmental parameters, for example, the spring constant $k_0$ and total wire mass $m$, are varied between the training and test scenarios.

As an example of disturbance caused by an adversary, we assume that the adversary can affect the wind speed in the simulation.
Thus, the adversary can append discontinuous additional wind to continuous wind in the environment.
At every time step $t$ in the training scenario, by considering the wind speed in the environment $\bm v_\mathrm e^{(k)}\in\mathbb R^3$ and the additional wind speed appended by the adversary $\bm v_\mathrm a^{(k)}\in\mathbb R^3$, the wind speed $\bm v_\mathrm o^{(k)}\in\mathbb R^3$ in \eqref{eq:pos_vel} is calculated as
\begin{align}
  \bm v_\mathrm o^{(k)} = \bm v_\mathrm e^{(k)} + \bm v_\mathrm a^{(k)}.
\end{align}
Conversely, because the adversary does not exist in the test scenarios, the wind speed $\bm v_\mathrm o^{(k)}\in\mathbb R^3$ in \eqref{eq:pos_vel} is given by
\begin{align}
  \bm v_\mathrm o^{(k)} = \bm v_\mathrm e^{(k)}.
\end{align}

\subsection{State, Action, and Reward}
The state set $\mathcal S$ of the protagonist and adversary is defined as
\begin{align}
  \label{eq:state_space}
  \mathcal S \coloneqq \mathcal S_{\bm x}\times\mathcal S_{\bm v}\times\mathcal S_{\bm b},
\end{align}
where $\mathcal S_{\bm x}\coloneqq\{\,\bm x\mid\bm x\in\mathbb R^3\,\}$ and $\mathcal S_{\bm v}\coloneqq\{\,\bm v\mid\bm v\in\mathbb R^3\,\}$ denote the set of possible three-dimensional positions and velocities of the SBS, respectively.
Note that for the practical implementation, these three-dimensional positions/velocities can be obtained if the SBS involves an accelerometer and then integrates the measured accelerations.
This position/velocity acquisition involves an error, and this error may affect the accuracy of the beam-tracking.
Nonetheless, we assume that these three-dimensional positions/velocities can be obtained without any measurement errors in view of the scope of this study.
As discussed in Section~I, our main scope is to validate the effectiveness of adding an adversary during training, thereby confirming the feasibility of our concept of zero-shot adaptation.
This objective can be achieved by comparing the proposed method (with the adversary) with the baseline method (without the adversary), and the measurement error does not have a specific impact on this comparison because the inaccuracy of the beam-tracking due to the measurement error occurs commonly in both methods.
Hence, this assumption is sufficient to validate the contribution of this study, and delving into the measurement precisions of the positions/velocities is beyond the scope of this study.

In \eqref{eq:state_space}, $\mathcal S_{\bm b}\coloneqq\{\,\bm b\mid\bm b\in\mathbb R^3\,\}$ denotes the set of possible beam directions, where beam direction at time step $k$ $\bm b^{(k)}$ is given by
\begin{align}
  \bm b^{(k)} = \big[\sin\theta^{(k)}_\mathrm s\cos\phi^{(k)}_\mathrm s, \sin\theta^{(k)}_\mathrm s\sin\phi^{(k)}_\mathrm s, \cos\theta^{(k)}_\mathrm s\big]^{\mathrm{T}}.
\end{align}
These state settings are consistent with our previous works \cite{shinzaki2020deep, koda2020millimeter} to ensure a fair comparison.

The action set of the protagonist $\mathcal A_\mathrm p$ is defined as
\begin{align}
  \mathcal A_\mathrm p \coloneqq \{\mathrm{stay}, \mathrm{up},\mathrm{down},\mathrm{left},\mathrm{right}\},
\end{align}
where the action $\mathrm{stay}$ denotes that the beam direction is maintained.
Moreover, the actions $\mathrm{up}$, $\mathrm{down}$, $\mathrm{left}$, and $\mathrm{right}$ denote that the beam direction is moved up, down, left, and right by $\beta$, respectively.
The actions for the zenith and azimuth angle $a^{(k)}_\theta, a^{(k)}_\phi$ are given by
\begin{align}
  \left[a^{(k)}_\theta, a^{(k)}_\phi\right] =
  \begin{cases}
    [0, 0]^{\mathrm{T}},  & a_k = \mathrm{stay};  \\
    [-1, 0]^{\mathrm{T}}, & a_k = \mathrm{up};    \\
    [1, 0]^{\mathrm{T}},  & a_k = \mathrm{down};  \\
    [0, 1]^{\mathrm{T}},  & a_k = \mathrm{left};  \\
    [0, -1]^{\mathrm{T}}, & a_k = \mathrm{right}.
  \end{cases}
\end{align}

The action set of the adversary $\mathcal A_\mathrm a$ is defined as
\begin{align}
  \mathcal A_\mathrm a \coloneqq \{\mathrm{stay}, \mathrm{up},\mathrm{down},\mathrm{left},\mathrm{right}, \mathrm{front}, \mathrm{back}\},
\end{align}
where the action $\mathrm{stay}$ denotes that no additional wind is appended.
Furthermore, in the training scenarios, the actions $\mathrm{up}$, $\mathrm{down}$, $\mathrm{left}$, $\mathrm{right}$, $\mathrm{front}$, and $\mathrm{back}$ denote wind in the upward, downward, leftward, rightward, forward, and backward directions with a wind speed of $v_\mathrm a$.
The additional wind speed by the adversary $\bm v_\mathrm a^{(k)}$ is given by
\begin{align}
  \bm v_\mathrm a^{(k)} =
  \begin{cases}
    [0, 0, 0]^{\mathrm{T}},            & a_k = \mathrm{stay};  \\
    [0, 0, v_\mathrm a]^{\mathrm{T}},  & a_k = \mathrm{up};    \\
    [0, 0, -v_\mathrm a]^{\mathrm{T}}, & a_k = \mathrm{down};  \\
    [-v_\mathrm a, 0, 0]^{\mathrm{T}}, & a_k = \mathrm{left};  \\
    [v_\mathrm a, 0, 0]^{\mathrm{T}},  & a_k = \mathrm{right}; \\
    [0, v_\mathrm a, 0]^{\mathrm{T}},  & a_k = \mathrm{front}; \\
    [0, -v_\mathrm a, 0]^{\mathrm{T}}, & a_k = \mathrm{back}.
  \end{cases}
\end{align}

The immediate reward is defined as the instantaneous received signal power at the next step, which is clipped using the technique in \cite{mnih2015human}:
\begin{align}
  \label{eq:clipping}
  r_{\mathrm p, k} =
  \begin{cases}
    1,                                             & (P^{(k+1)}_\mathrm r-b_\mathrm c)/d_\mathrm c >1;   \\
    (P^{(k+1)}_\mathrm r-b_\mathrm c)/d_\mathrm c, & -1\leq(P-b_\mathrm c)/d_\mathrm c\leq 1;            \\
    -1,                                            & (P^{(k+1)}_\mathrm r-b_\mathrm c)/d_\mathrm c < -1,
  \end{cases}
\end{align}
where $P^{(k+1)}_\mathrm r \coloneqq P_\mathrm{r}\left(d^{(k+1)}, \theta^{(k+1)}, \phi^{(k+1)}, \theta_\mathrm s^{(k+1)}, \phi_\mathrm s^{(k+1)}\right)$.
Moreover, $b_\mathrm c$ and $d_\mathrm c$ denote the offset and scale of the clipping, respectively.
The immediate reward of the adversary $r_{\mathrm a, k}$ is defined by inverting the sign of that of the protagonist  to encourage the adversary to disturb the beam-tracking agent, that is, $r_{\mathrm a, k} = -r_{\mathrm p, k}$.

\subsection{Adversarial RL Algorithm}
\label{subsec:training_procedure}
Given the definition of the state, action, and reward, the training of the protagonist beam-tracking agent and adversary is performed via deep Q-learning \cite{mnih2015human} for the two agents.
In deep Q-learning, the following value, termed the optimal action-value function, is predicted via a neural network:
\begin{multline}
  Q_{\mathit{agent}}^{\star}(s, a) = \mathbb{E}_{\pi_{\mathit{agent}}^{\star}}\!\left[\sum_{k'=0}^\infty \gamma^{k'} r_{\mathit{agent}, k+1+k'}\,\middle|\,s_k = s, a_{\mathit{agent}, k} = a\right],\\
  s\in\mathcal{S}, a\in\mathcal{A}_{\mathit{agent}},
\end{multline}
where $s_k$ and $\gamma\in[0, 1)$ denote the state at the time step $k$ and discount factor, respectively.
For $\mathit{agent}\in\{\mathrm{p}, \mathrm{a}\}$, which indicates the protagonist or adversary, $a_{\mathit{agent}, k}$ and $r_{\mathit{agent}, k}$ denote the action and reward given to $\mathit{agent}$ at time step $k$, respectively.
The term $\pi_{\mathit{agent}}^{\star}:\mathcal{S}\to\mathcal{A}_{\mathit{agent}}$ is the optimal policy, which means the action rule to maximize the discounted reward.
Although finding the optimal policy is the main objective of this algorithm, in deep Q-learning, we can find the optimal action value function first, and then we can determine the optimal policy by: $\textstyle \argmax_{a\in\mathcal{A}_{\mathit{agent}}}Q_{\mathit{agent}}^{\star}(s, a)$.
Hence, the problem boils down to finding the optimal action-value function, which is conducted by training a neural network known as a deep Q network (DQN) in deep Q-learning such that the DQN is a good approximation of the optimal action-value function.
Let $Q_{\mathit{agent}}(s, a; \bm \theta_{\mathit{agent}})$ denote the DQN for $\mathit{agent}$.

The procedure used to train the two DQNs is detailed below.
Note that procedures 1 and 2 are conducted in every time step, whereas procedures 3, 4, 5 are performed on a per-episode basis.
Here, we let the episode be the finite time steps for $k = 1, 2, \dots, \lfloor T / \tau \rfloor$.

\vspace{.3em}\noindent
\textbf{1. Exploring and Storing Experience.}\quad
In this procedure, the protagonist and adversary collect the ingredients to create training data, termed experience.
The experience is defined as $(s_k, a_{\mathit{agent}, k}, r_{\mathit{agent}, k}, s_{k + 1})$ for $\mathit{agent}$ and is collected while interacting with the environment, that is, taking the action, obtaining the rewards, and the subsequent states.
In this procedure, the protagonist and adversary synchronously perform the action following $\epsilon$-greedy policies, which is a general assumption that includes the prior work for adversarial RL\cite{pinto2017robust}.
The experience for $\mathit{agent}$ is stored in the experience memory denoted by $\mathcal{D}_{\mathit{agent}}$.

\vspace{.3em}\noindent
\textbf{2. Training DQNs.}\quad
Given the experience memories $\mathcal{D}_{\mathit{agent}}$ for $\mathit{agent}\in\{\mathrm{p}, \mathrm{a}\}$, the DQNs are trained.
In deep Q-learning, the DQN is trained to minimize the difference metric between $Q_{\mathit{agent}}(s, a; \bm \theta_{\mathit{agent}})$ and $\mathit{target}_{\mathit{agent}}\coloneqq r + \gamma \max_{a\in\mathcal{A}_{\mathit{agent}}}Q_{\mathit{agent}}(s, a; \bm \theta^{-}_{\mathit{agent}})$, where $Q_{\mathit{agent}}(s, a; \bm \theta^{-}_{\mathit{agent}})$ is termed the target network and is updated less frequently than the DQN.
We refrain from delve into the details of the target network to enable us to focus on the training procedure. Interested readers are encouraged to refer to the paper of Mnih \textit{et al.}\cite{mnih2015human}.

The training procedure is as follows.
First, we calculate $\mathit{target}_{\mathit{agent}}$ by sampling the experience from $\mathcal{D}_{\mathit{agent}}$ uniformly.
Subsequently, we train each DQN to minimize the difference metric between the DQN $Q_{\mathit{agent}}(s, a; \bm \theta_{\mathit{agent}})$ and $\mathit{target}_{\mathit{agent}}$ via the Adam optimizer \cite{kingma2015adam}.
Note that the protagonist and adversary DQNs are trained synchronously, which is found to be sufficient to demonstrate the robustness of the protagonist against training-test gaps.

During this training procedure, we leveraged more advanced techniques in the evaluation described in Section~\ref{sec:simulation}.
More specifically, we leveraged the Huber loss \cite{varga2018deeprn} as the difference metric instead of taking the square of $Q_{\mathit{agent}}(s, a; \bm \theta_{\mathit{agent}})-\mathit{target}_{\mathit{agent}}$.
In addition, we leveraged dueling DQN \cite{wang2015dueling} in the experiment.
However, detailing these techniques is beyond the objective of this section; hence, we detailed these techniques in the Appendix.

\vspace{.3em}\noindent
\textbf{3. Updating Target DQNs.}\quad
The parameters of the target DQNs $\bm \theta^{-}_{\mathit{agent}}$ are updated such that $\bm \theta^{-}_{\mathit{agent}}$ is equal to $\bm \theta_{\mathit{agent}}$.
This update is generally performed less frequently than that of the update the main DQNs\cite{mnih2015human}; hence, we conduct this procedure when every $C$ episode elapses.
This target network update is performed synchronously for the protagonist and adversary.

\vspace{.3em}\noindent
\textbf{4. Checking Protagonist Performance.}\quad  This procedure is performed to check whether an appropriate beam-tracking policy is learned regardless of the adversary.
In this procedure, for additional $\mathit{test}$ steps, the protagonist performs beam-tracking by greedily determining the action with respect to the DQN $Q_{\mathrm{p}}(s, a;\bm \theta_{\mathrm{p}})$, that is, it chooses the action by: $\textstyle \argmax_{a\in\mathcal{A}_{\mathit{p}}}Q_{\mathrm{p}}^{\star}(s, a;\bm \theta_{\mathrm{p}})$, and the received power averaged for the steps is obtained.
Therein, the adversary is not activated to check the protagonist performance in view of real deployments.

\vspace{.3em}\noindent
  \textbf{5. Checking Adversary Performance.}\quad This procedure is performed to check whether the adversary can with certainty learn a policy to disturb a protagonist.
  In this procedure, for additional $\mathit{test}$ steps, the adversary disturbs the protagonist by greedily determining the action with respect to the DQN $Q_{\mathrm{a}}(s, a;\bm \theta_{\mathrm{a}})$.
  In contrast to the previous procedure, checking the performance of adversary still requires the existence of a protagonist, whereas using the protagonist in this learning procedure $Q_{\mathrm{p}}(s, a;\bm \theta_{\mathrm{p}})$ underestimates the performance of the adversary because the protagonist may already be robust against the adversary.
  To avoid this, we prepared another proxy protagonist, \textit{pre-trained} without the adversary, and it is this proxy protagonist that is disturbed by the adversary in this phase.
  This proxy protagonist did not learn a robust policy against the adversary; therefore, we can keep track of the performances of the adversary without underestimation.
  Note that by letting $Q_{\mathrm{p}}(s, a;\bm \theta_{\mathrm{p, proxy}})$ denote the DQN of the proxy protagonist, this progagonist also determines the action greedily with respect to $Q_{\mathrm{p}}(s, a;\bm \theta_{\mathrm{p, proxy}})$.
  The performance of the adversary is measured by the average received power that is obtained in the proxy protagoist during this phase.

\vspace{.3em}
  These procedures are iterated for a predefined number of episodes $M$.
  This value of $M$ is determined to be sufficiently longer than the convergence of the protagonist performance, which could be measured in procedure~4.
   Note that during the training procedure, the adversary is always active except for procedure~4.
  The overall training procedure is summarized in Algorithm~\ref{alg:adversarial}.

\begin{algorithm}[t]
  \caption{Training the protagonist beam-tracking agent and adversary via adversarial RL}
  \label{alg:adversarial}
  \setlength{\baselineskip}{11.4pt}
  \begin{algorithmic}
    \State Initialize main DQN and target DQN of protagonist, i.e., $Q_\mathrm p(s,a;\bm\theta_\mathrm p)$ and $Q_\mathrm p(s,a;\bm\theta^-_\mathrm p)$, respectively, and experience memory of protagonist $\mathcal D_\mathrm p$
    \State Initialize main DQN and target DQN of adversary, i.e., $Q_\mathrm a(s,a;\bm\theta_\mathrm a)$ and $Q_\mathrm a(s,a;\bm\theta^-_\mathrm a)$, respectively, experience memory of adversary $\mathcal D_\mathrm a$
    \State \parbox[t]{205pt}{
      Observe initial state $s_{1}$ by obtaining $\bm x_\mathrm S^{(1)}$, velocity $\bm v_\mathrm S^{(1)}$, and beam direction $\bm b_\mathrm S^{(1)}$ \strut
    }
    \For{Episode $e = 1, 2, \dots, M$}
    \For{time step $k = 1, 2,\dots,\lfloor T/\tau \rfloor$}
    \Exploration
    \State \parbox[t]{190pt}{
      Select actions of protagonist and adversary  $a_{\mathrm{p},k}$ and $a_{\mathrm{a}, k}$ synchronously
    }
    \State Calculate reward for protagonist $r_{k,\mathrm{p}}$  from \eqref{eq:clipping}
    \State Calculate reward $r_{k,\mathrm{a}}$ by $r_{k,\mathrm{a}}\leftarrow -r_{k,\mathrm{p}}$
    \State \parbox[t]{205pt}{
      Observe state $s_{k+1}$ by obtaining $\bm x_\mathrm S^{(k+1)}$, velocity $\bm v_\mathrm S^{(k+1)}$, and beam direction $\bm b_\mathrm S^{(k+1)}$ \strut
    }
    \State Store experience $(s_{k}, a_{\mathrm p,k}, r_{\mathrm p,k}, s_{k+1})$ into $\mathcal D_\mathrm p$
    \State Store experience $(s_{k}, a_{\mathrm a,k}, r_{\mathrm a,k}, s_{k+1})$ into $\mathcal D_\mathrm a$
    \Training
    \State \parbox[t]{220pt}{Synchronously update weights of main DQNs of protagonist and adversary $\bm \theta_\mathrm p, \bm \theta_\mathrm a$\strut}
    \Target
    \If{$e\equiv 0\ (\mathrm{mod}\,C)$}
    \State Update target DQNs $\bm\theta^-_\mathrm p\gets\bm\theta_\mathrm p$ and $\bm\theta^-_\mathrm a\gets\bm\theta_\mathrm a$
    \EndIf
    \EndFor
    \PerformanceCheck
    \For{time step $k = \lfloor T/\tau \rfloor + 1,\dots,\lfloor T/\tau \rfloor + \mathit{test}$}
    \State \parbox[t]{190pt}{Select actions of protagonist $a_{\mathrm{p},k}$ greedily w.r.t. $Q_\mathrm{p}(s, a; \theta_{\mathrm{p}})$}
    \State \parbox[t]{205pt}{
      Observe state $s_{k+1}$ in the same way as that in procedure~1 \strut
    }
    \State Observe received power
    \EndFor
    \PerformanceCheckad
    \For{time step $k = \lfloor T/\tau \rfloor + 1,\dots,\lfloor T/\tau \rfloor + \mathit{test}$}
    \State \parbox[t]{190pt}{Select actions of proxy protagonist $a_{\mathrm{p},k}$ greedily w.r.t. $Q_\mathrm{p}(s, a; \theta_{\mathrm{p, proxy}})$ (pre-trained without adversary)}
    \State \parbox[t]{190pt}{Select actions of adversary $a_{\mathrm{a},k}$ greedily w.r.t. $Q_\mathrm{a}(s, a; \theta_{\mathrm{a}})$}
    \State \parbox[t]{205pt}{
      Observe state $s_{k+1}$ in the same way as that in procedure~1 \strut
    }
    \State Observe received power
    \EndFor
    \State Check average received power
    \EndFor
  \end{algorithmic}
\end{algorithm}

\section{Simulation Results}
\label{sec:simulation}
\subsection{Simulation Parameters}
The simulation parameters are listed in Table~\ref{tbl:SimurationParam}, where $\bm E\in\mathbb R^{3\times 3}$ is an identity matrix.
The SBS is installed at the midpoint of the overhead messenger wire, which experiences significant movement as a result of the wind.
The wind speed in the environment $\bm v_\mathrm e^{(k)}$ is given by
\begin{align}
  \bm v_\mathrm e^{(k)} = \left[5\sin\frac{2\pi k\tau}{4}, 5\sin\frac{2\pi k\tau}{6}, 5\sin\frac{2\pi k\tau}{8}\right]^{\mathrm{T}},
\end{align}
which follows our previous studies\cite{shinzaki2020deep, koda2020millimeter}.

\subsubsection{Antenna Pattern}
\begin{figure}[t]
  \centering
  \subfigure[{$\phi\in[-180^\circ,180^\circ]$.}]{
    \includegraphics[width=0.45\columnwidth]{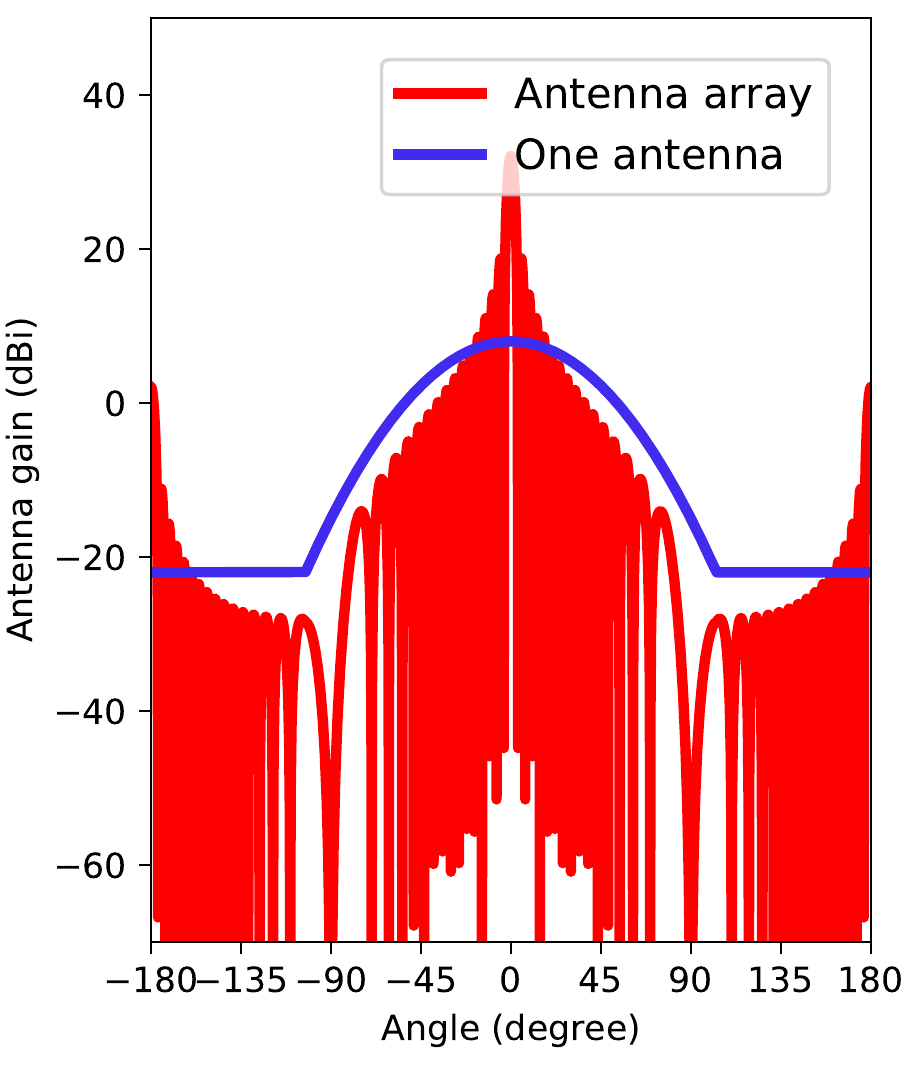}
    \label{fig:gcn_layer}
  }
  \centering
  \subfigure[{$\phi\in[-8^\circ,8^\circ]$.}]{
    \includegraphics[width=0.45\columnwidth]{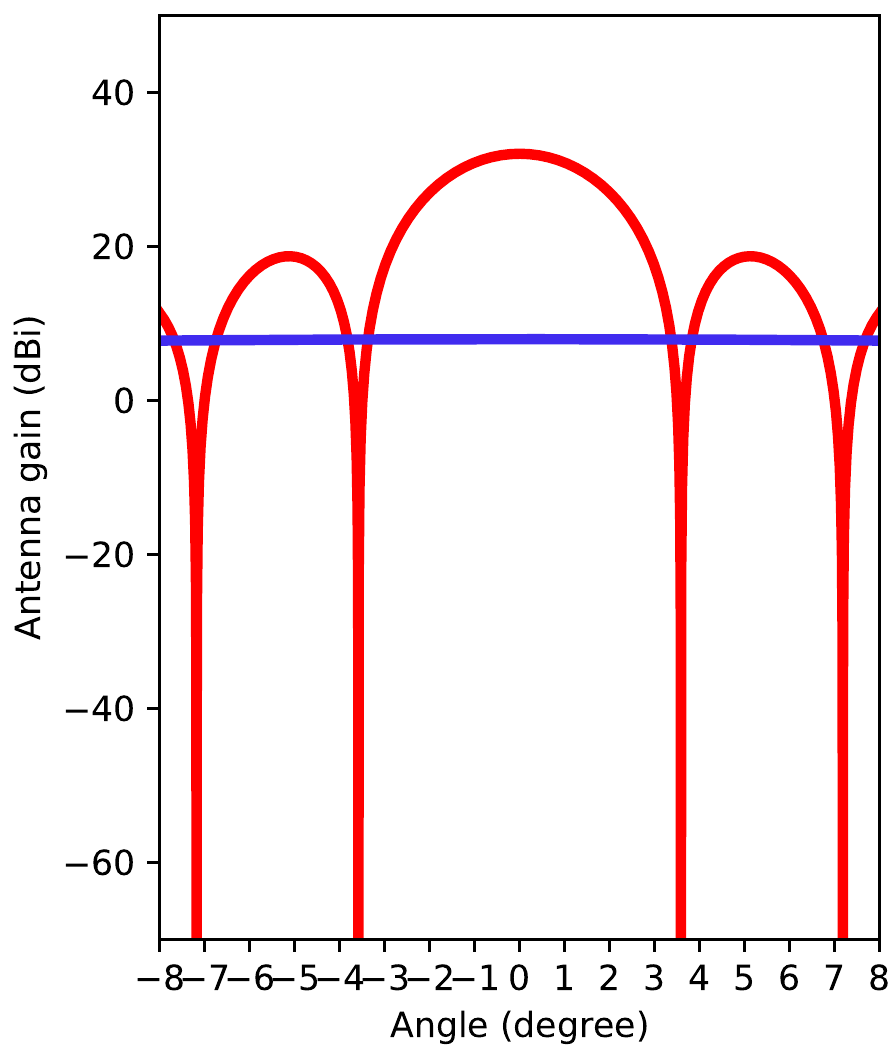}
    \label{fig:nn_layer}
  }
  \caption{Antenna pattern. The vertical angle $\theta$ was fixed at $90^\circ$.}
  \label{fig:antenna_pattern}
\end{figure}

Fig.~\ref{fig:antenna_pattern} shows the antenna pattern based on the simulation parameters listed in Table~\ref{tbl:SimurationParam}.
The red trace represents the transmission antenna gain with the array factor $\mathit{AF}(\theta, \phi, \theta_\mathrm s, \phi_\mathrm s)$.
The blue curve represents the transmission antenna gain without the array factor, that is, $\mathit{AF}(\theta, \phi, \theta_\mathrm s, \phi_\mathrm s) = 0$ in \eqref{eq:antenna_gain}.
The directivity of the transmission antenna gain with the array factor is higher than that of the transmission antenna gain without the array factor.
In the transmission antenna gain with the array factor, $3.6^\circ$ is the positive minimum value of the local minimum values.

\subsubsection{Architecture of the Neural Network}
\begin{figure}[t]
  \centering
  \includegraphics[width=0.8\columnwidth]{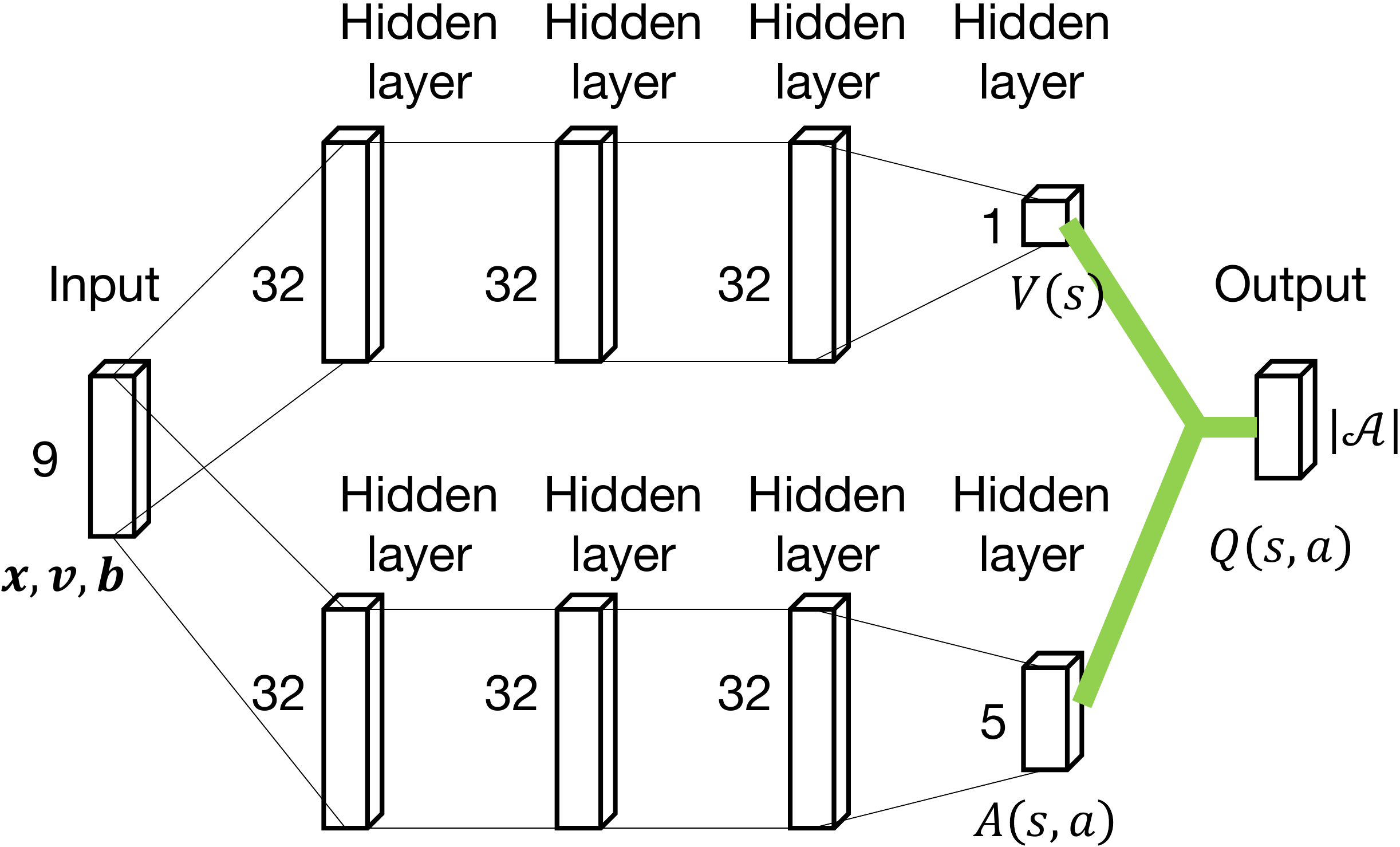}
  \caption{Architecture of the NN of the protagonist and adversary.}
  \label{fig:nn}
\end{figure}

We used a neural network with four hidden layers, as shown in Fig.~\ref{fig:nn}, where $|\mathcal A|$ denotes the number of actions.
The number of actions of the protagonist $|\mathcal A_\mathrm p| = 5$ and the number of actions of the adversary $|\mathcal A_\mathrm a| = 7$.
The hidden layers were all fully connected and had $32$ units.
The activation function of the hidden layers is the rectified linear unit $R(x)$, which is given by
\begin{align}
  R(x) = \max \{x, 0\}.
\end{align}
Adam \cite{kingma2015adam} was used as the gradient descent method, and the learning rates of the protagonist and adversary $\alpha_\mathrm p, \alpha_\mathrm a$ were $10^{-3}$.

\begin{table}[t]
  \centering
  \caption{Simulation Parameters}
  \begin{tabular}{cc}
    \toprule
    Height of endpoints of wire $h_\mathrm w$                   & $5\,\mathrm{m}$                  \\
    Distance between endpoints $d_\mathrm w$                    & $10\,\mathrm{m}$                 \\
    Height of gateway BS $h_\mathrm r$                          & $5\,\mathrm{m}$                  \\
    Distance between wire and gateway BS $d_\mathrm r$          & $5\,\mathrm{m}$                  \\
    Transmission power $P_\mathrm t$                            & $23\,\mathrm{dBm}$               \\
    Radio-wave wavelength $\lambda$                             & $5\,\mathrm{mm}$                 \\
    Receiver antenna gain $A_\mathrm r$                         & $8\,\mathrm{dBi}$                \\
    Gravitational acceleration $\bm{g}$                         & $[0, 0, -9.8]\,\mathrm{ms^{-2}}$ \\
    Spring constant $k_0$                                       & $100\,\mathrm{N}\mathrm{m}^{-1}$ \\
    Drag constant $c_0$                                         & $1\,\mathrm{s}^{-1}$             \\
    Number of points $N$                                        & 11                               \\
    Total wire mass $m$                                         & $10\,\mathrm{kg}$                \\
    Covariance matrix of the wind speed $\bm{V}_0$              & $0.1\bm{E}$                      \\
    Vertical $3\,\mathrm{dB}$ beamwidth $\theta_{3\mathrm{dB}}$ & $65^\circ$                       \\
    Horizontal $3\,\mathrm{dB}$ beamwidth $\phi_{3\mathrm{dB}}$ & $65^\circ$                       \\
    Side-lobe level limit $\mathit{SLA}_\mathrm V$              & $30\,\mathrm{dB}$                \\
    Front-back ratio $A_\mathrm m$                              & $30\,\mathrm{dB}$                \\
    Number of vertical elements $n_\mathrm V$                   & 32                               \\
    Number of horizontal elements $n_\mathrm H$                 & 32                               \\
    Zenith steering angle $\theta_\mathrm s$                    & $90^\circ$                       \\
    Azimuth steering angle $\phi_\mathrm s$                     & $0^\circ$                        \\
    Vertical spacing distance $\Delta_\mathrm V$                & $2.5\,\mathrm{mm}$               \\
    Horizontal spacing distance $\Delta_\mathrm H$              & $2.5\,\mathrm{mm}$               \\
    Number of episodes $M$                                      & $400$                            \\
    Point of SBS installation                                   & Point 6                          \\
    Observation time $T$                                        & $10\,\mathrm{s}$                 \\
    Observation interval $\tau$                                 & $0.01\,\mathrm{s}$               \\
    Angle when moving beam direction $\beta$                    & $1^\circ$                        \\
    Constant in \eqref{eq:clipping} $b_\mathrm c$               & 27                               \\
    Constant in \eqref{eq:clipping} $d_\mathrm c$               & 3                                \\
    Exploration rate for $\epsilon$-greedy policy               & 0.2                              \\
    Discount factor $\gamma$                                    & $0.99$                           \\
    Target DQN frequency $C$                                    & $5$                              \\
    Wind speed added by the adversary $v_\mathrm a$             & $10\,\mathrm{m/s}$               \\
    \bottomrule
  \end{tabular}
  \label{tbl:SimurationParam}
\end{table}

\subsection{Baseline Method}
To evaluate the average received signal power of the learned policy of the protagonist, we compared the proposed method with the \textit{stay method} and the \textit{upper-limit method}.
Because these compared methods aim to evaluate the policy of the protagonist without the disturbance caused by the adversary, the adversary does not exist in the stay or upper-limit methods.
In the stay method, the beam direction is fixed in the initial beam direction.
The policy $\pi(a\,\vert\,s)$ of the stay method is given by
\begin{align}
  \pi(a\,\vert\,s) =
  \begin{cases}
    1, & a = \mathrm{stay};                                          \\
    0, & a = \mathrm{up},\mathrm{down},\mathrm{left},\mathrm{right}.
  \end{cases}
\end{align}
In the upper-limit method, the actions maximize the transmission antenna gain such that the superior performance is delivered in terms of the received signal power.

To evaluate the robustness of the learned policy of the protagonist, we compared the proposed method with the \textit{no adversary method} and \textit{random 10\,m/s method}.
In the no adversary method, the policy of the protagonist is learned in the scenario in which the adversary appends no additional wind, i.e., $v_\mathrm a=0$.
In the random $10\,\mathrm{m/s}$ method, the policy of the protagonist is learned in the scenario in which the adversary takes each action at random with equal probability.

\subsection{Learning Curve}
\begin{figure}[t]
  \centering
  \includegraphics[width=0.8\columnwidth]{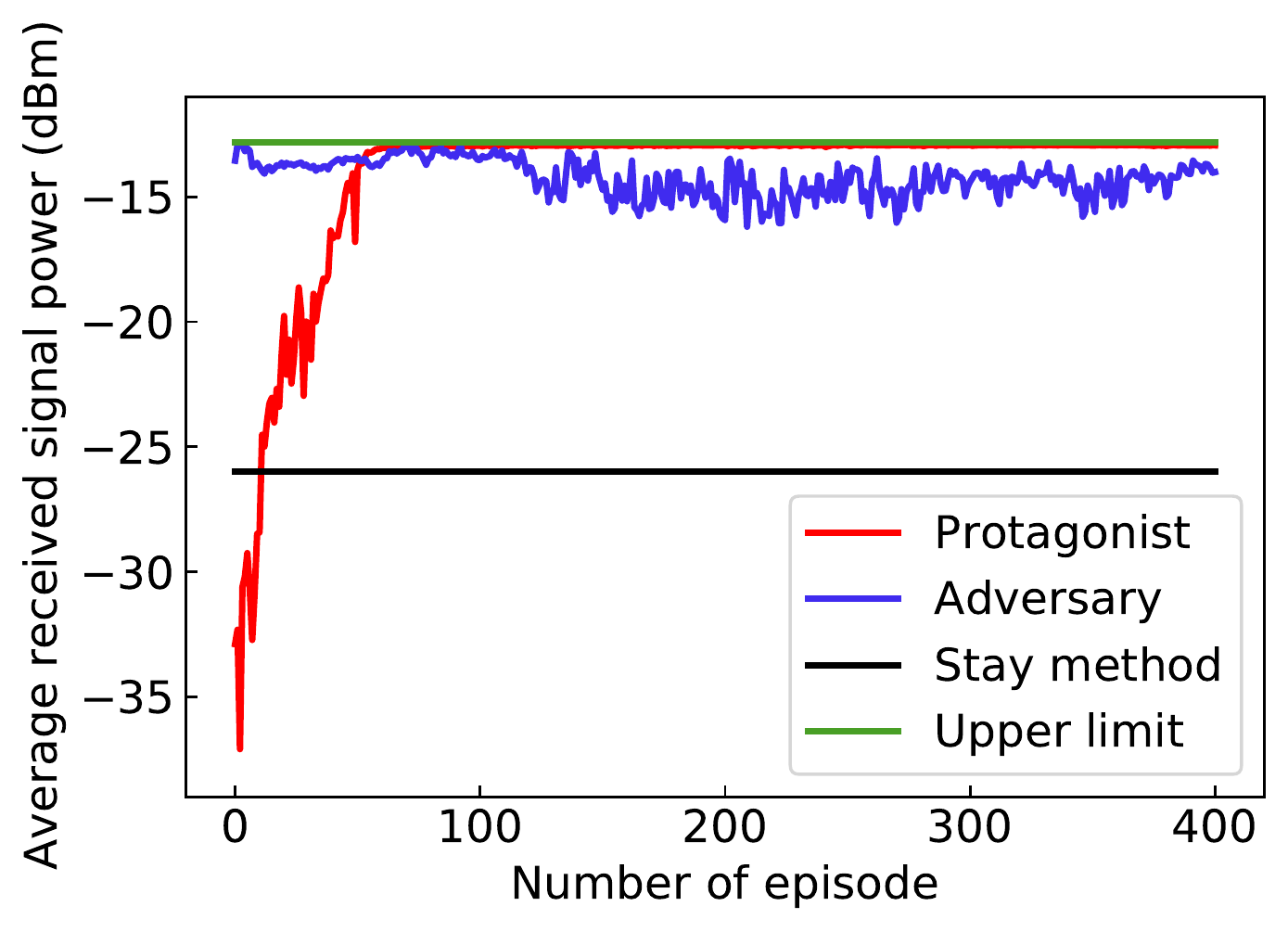}
  \caption{Learning curve of the protagonist and the adversary. Only in the adversary performance in the blue curve, the lower received power is better. As the episode elapses, the performance of both the protagonist and adversary increases as the episode elapses.}
  \label{fig:adversarial_learning_curve_10}
\end{figure}

The result of the performance check of the protagonist and adversary during training is shown in Fig.~\ref{fig:adversarial_learning_curve_10}, along with the performance of the baseline methods.
  The red curve shows the performance of the protagonist obtained during the procedure ``4.~Checking Protagonist Performance'' whereas the blue curve shows the performance of the adversary obtained during the procedure ``5.~Checking Adversary Performance,'' which is detailed in Section~\ref{subsec:training_procedure}.
  Note that only with respect to the performance of the adversary (represented by the blue curve), lower received power is better because the performance of the adversary should be measured with the extent to which it disturbs the proxy protagonist.
  As shown in Fig.~\ref{fig:adversarial_learning_curve_10}, regardless of the adversarial training, the performance of both the protagonist and adversary increases as the episode elapses.
  This first shows the feasibility of the protagonist achieving an appropriate beam-tracking policy that converges closely to the upper limit regardless of the adversary.
  Moreover, although the received power fluctuates, the adversary can basically obtain the policy of disturbing the protagonist as the episode elapses.

\subsection{Robustness of the Learned Policy}
\label{subsec:simu_robustness}
\begin{figure}[t]
  \centering
  \includegraphics[width=0.8\columnwidth]{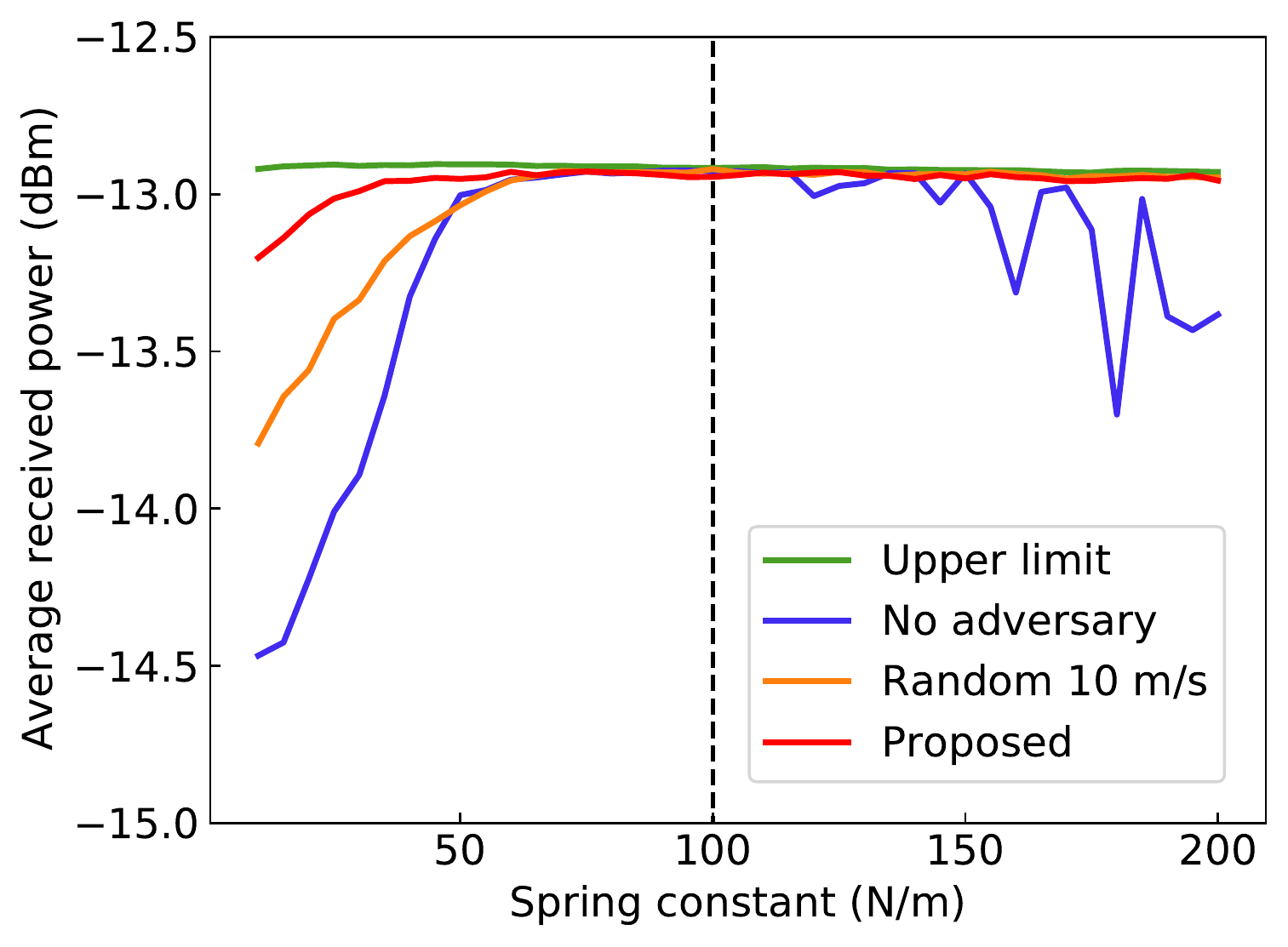}
  \caption{Robustness to variations in the spring constant of the overhead messenger wire. In the test scenarios, the average received signal power of the proposed method is more robust to different spring constants than the compared methods. The dashed line represents the training parameter of the spring constant.}
  \label{fig:spring_constant}
\end{figure}

To demonstrate the robustness to variations in the spring constant, we compared the average received signal power of the proposed method with that of the no adversary and random $10\,\mathrm{m/s}$ methods, as shown in Fig.~\ref{fig:spring_constant}.
In Fig.~\ref{fig:spring_constant}, the spring constant of the messenger wire is shown on the horizontal axis, where the training parameter of the spring constant is represented by the dashed line.
When the spring constant $k_0=100\,\mathrm{N}\mathrm{m}^{-1}$, which is the setting in the training, all the methods, including baseline methods, exhibited an almost identical amount of  received power, that is, approximately $-12.9\,\mathrm{dBm}$.
However, in the scenario of when the spring constant was lower than that in training, that is, $k_0=10\,\mathrm{Nm^{-1}}$, the average received signal power of the proposed method was $-13.2\,\mathrm{dBm}$, whereas that of the no adversary and random $10\,\mathrm{m/s}$ methods dropped to $-14.5$ and $-13.8\,\mathrm{dBm}$, respectively.
Indeed, the random 10\,m/s method (orange curve) is advantageous to enable the protagonist to learn a robust policy when compared to no adversary method.
However, in contrast to the random 10\,m/s method, the proposed method with the adversary is more effective for the protagonist to learn a robust policy.

\begin{figure}[t]
  \centering
  \includegraphics[width=0.8\columnwidth]{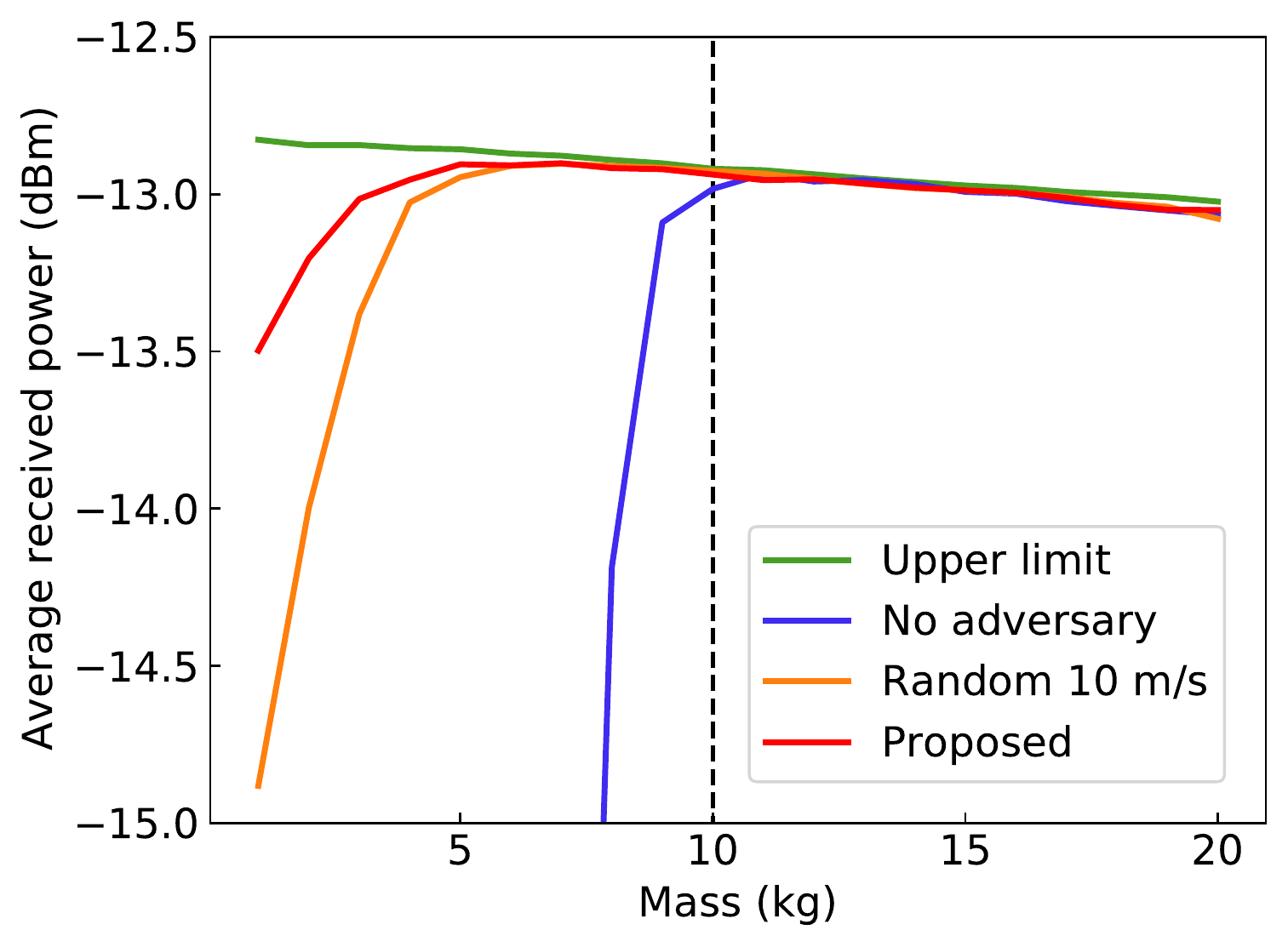}
  \caption{Robustness to variations in total wire mass. In the test scenarios, the average received signal power of the proposed method is more robust to various total wire masses than the compared methods. The dashed line represents the training parameter of the total wire mass.}
  \label{fig:mass}
\end{figure}

To evaluate the robustness to the variations in the total wire mass, we compared the average received signal power of the proposed method with that of the no adversary and random $10\,\mathrm{m/s}$ methods, as shown in Fig.~\ref{fig:mass}.
In Fig.~\ref{fig:mass}, the total wire mass is shown on the horizontal axis, where the training parameter of the total wire mass is represented by the dashed line, that is, $10\,\mathrm{kg}$.
Similar to the results pertaining to the spring constant, the proposed methods exhibited a larger amount of received power for various amounts of the total wire mass.
Hence, the proposed method achieves more robust beam-tracking against the training-and-test gaps in terms of the total wire mass.
Note that these achievements are without any adaptive fine-tuning of the total wire mass and hence demonstrates the feasibility of the aforementioned zero-shot adaptation in mmWave beam-tracking.

\begin{figure}[t]
  \centering
  \begin{tabular}{c}
    \begin{minipage}[t]{0.384\hsize}
      \centering
      \includegraphics[width=1.0\columnwidth]{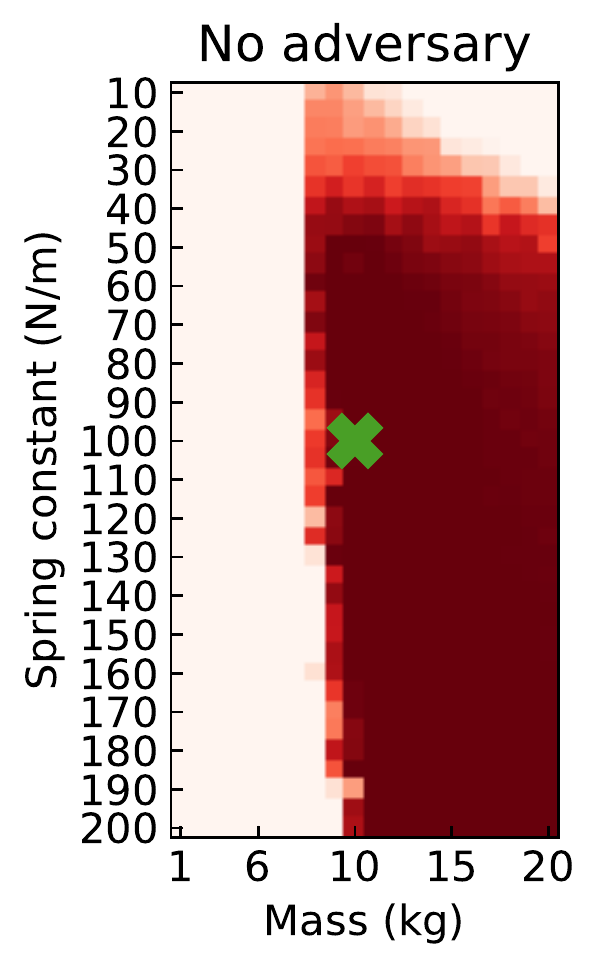}
    \end{minipage}
    \begin{minipage}[t]{0.57\hsize}
      \centering
      \includegraphics[width=1.0\columnwidth]{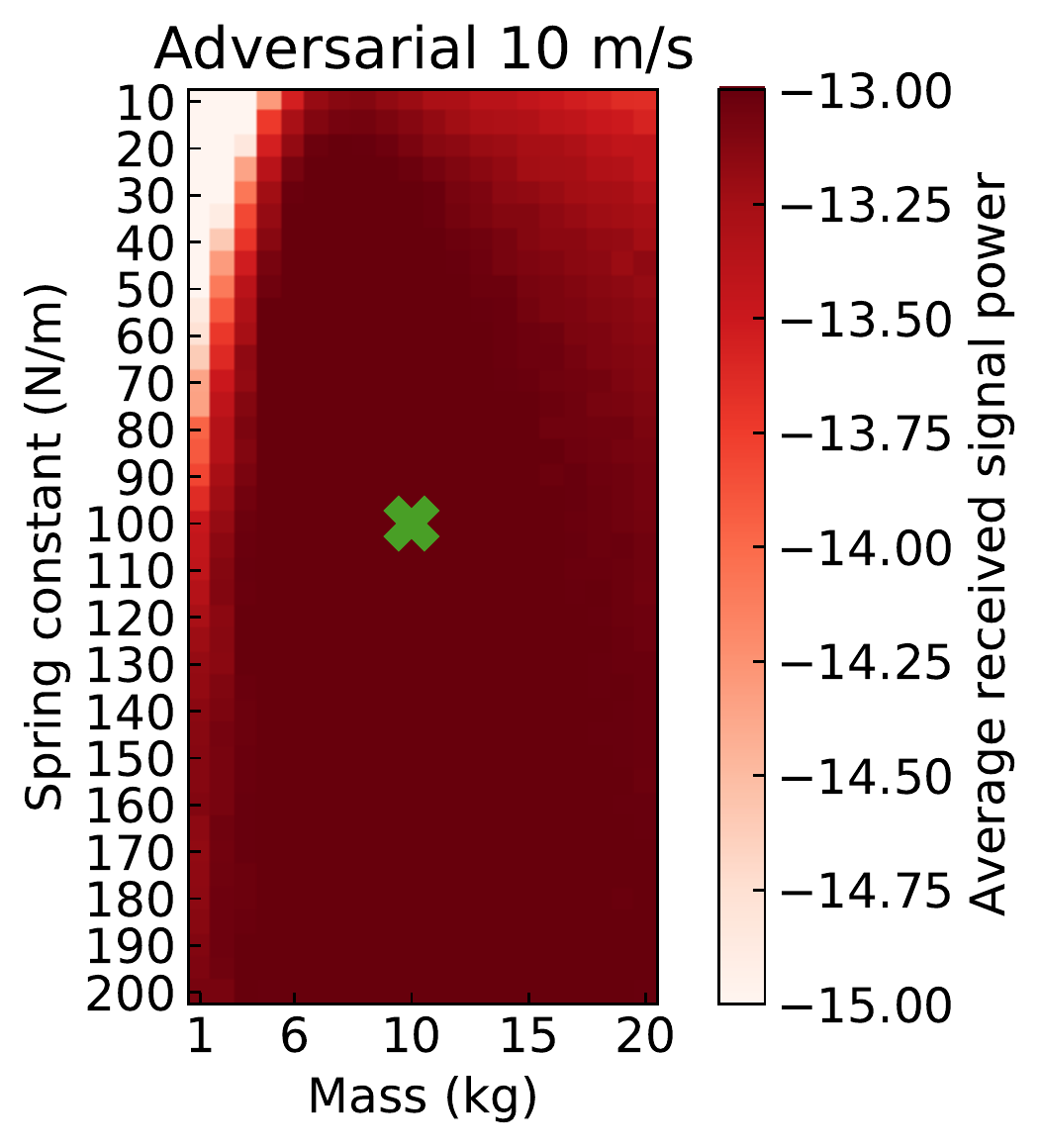}
    \end{minipage}
  \end{tabular}
  \caption{Heat map depicting the robustness to variations in the spring constant and the total wire mass of the messenger wire. In the test scenarios, the average received signal power of the proposed method is more robust to the variations in the total wire mass than the no adversary method. The cross mark represents the training parameters of the spring constant and total wire mass.}
  \label{fig:heatmap}
\end{figure}

We discuss the robustness to variations in the spring constants and total wire mass by plotting the average received signal power for various values of the spring constant and total wire mass in Fig.~\ref{fig:heatmap}.
The color of the heat map represents the average received signal power, where the cross mark represents the training parameters of the spring constant and total wire mass.
In Fig.~\ref{fig:heatmap}, the area in which the average amount of received power is higher, is shown in red, whereas that ohin which the average received power is low, is shown in white.
Fig.~\ref{fig:heatmap} demonstrates that the area representing the high average received power of the proposed method is wider than that of the no adversary method.
This implies that the proposed method enables the adaptation of various ranges of test settings for beam-tracking without the requirement for adaptive fine-tuning; thus, in this sense, this result demonstrates the feasibility of zero-shot adaptation in learning-based beam-tracking.

\section{Conclusion}
\label{sec:conclusion}
We discussed zero-shot adaptation in learning-based mmWave beam-tracking.
To demonstrate the feasibility of zero-shot adaptation, we proposed an adversarial RL-based beam-tracking method to obtain a robust beam-tracking policy to overcome the differences between the training and test scenarios, such as variations in the wire tension and the total wire mass.
We developed an RARL-based algorithm in which the adversary generated additional wind.
We demonstrated that the proposed method is more robust, not only than the no adversary method but also than the random $10\,\mathrm{m/s}$ method, when assuming an adversary without training.
This shows that disturbance by an intelligent adversary increased the robustness of the beam-tracking policy to variations in the wire tension, that is, the spring constant or total wire mass.
The improved robustness was greated than for an adversary without training.

\appendix

\vspace{.3em}\noindent\textbf{Huber Loss \cite{varga2018deeprn}.}\quad
The difference between $Q_{\mathit{agent}}(s, a; \bm \theta_{\mathit{agent}})$ is measured by the Huber loss \cite{varga2018deeprn}, which is given by:
\begin{align}
  \mathcal L(x) =
  \begin{cases}
    x^2/2,     & x \leq 1;         \\
    |x| - 0.5, & \text{otherwise},
  \end{cases}
\end{align}
where
\begin{align}
  x = r + \gamma \max_{a'}Q(s', a'; \theta^-) - Q(s,a;\bm\theta_i).
\end{align}


\vspace{.3em}\noindent\textbf{Dueling DQN (DDQN) \cite{wang2015dueling}.}\quad In dueling DQN, each Q network is divided into two networks.
More specifically, the action-value function $Q^\pi(s,a)$ is divided into a state value function $V^\pi(s)$ and an advantage function $A^\pi(s,a)$.
The state-value and advantage functions are expressed as follows:
\begin{align}
  V^\pi(s)   & = \mathbb E_{a\sim\pi(s)}[Q^\pi(s,a)], \notag \\
  A^\pi(s,a) & = Q^\pi(s,a) - V^\pi(s).
\end{align}
By using the state-value function approximated by a neural network $V(s;\bm\theta_V)$ and the advantage function approximated by NN $A(s,a';\bm\theta_A)$, the Q-network $Q(s,a;\bm\theta^-)$ is expressed as follows:
\begin{align}
  Q(s,a;\bm\theta) & = V(s;\bm\theta_V) \nonumber                                                                    \\
                   & \quad + \left(A(s,a;\bm\theta_A) - \frac{1}{|\mathcal A|} \sum_{a'} A(s,a';\bm\theta_A)\right),
\end{align}
where $\bm\theta_V$ and $\bm\theta_A$ are the weights used by neural network for approximating the state value and advantage functions, respectively.
Note that $\bm\theta=[\bm\theta_V, \bm\theta_A]$.

\bibliographystyle{IEEEtran}
\bibliography{IEEEabrv,main}
\end{document}